\pdfoutput=1

\documentclass[11pt]{article}

\usepackage[final]{acl}

\usepackage{times}
\usepackage{latexsym}
\usepackage{float}

\usepackage[T1]{fontenc}

\usepackage[utf8]{inputenc}

\usepackage{microtype}

\usepackage{inconsolata}

\usepackage{graphicx}

\usepackage{times}
\usepackage{soul}
\usepackage{url}
\usepackage[utf8]{inputenc}
\usepackage{amsmath}
\usepackage{amsthm}
\usepackage{booktabs}
\usepackage{algorithm}
\usepackage{algorithmic}
\usepackage[switch]{lineno}
\usepackage{tabularx}
\usepackage{booktabs}
\usepackage{bm}
\usepackage{bbm}
\usepackage{amssymb}
\usepackage{mathrsfs}
\usepackage{wrapfig}
\usepackage{subfig}
\usepackage{caption}
\usepackage{multicol}
\usepackage{multirow}
\usepackage{makecell}
\usepackage{enumerate}
\usepackage{threeparttable} 
\usepackage{tablefootnote} 

\title{Diversity-oriented Data Augmentation with Large Language Models}

\author{
 \textbf{Zaitian Wang\textsuperscript{1,2}},
 \textbf{Jinghan Zhang\textsuperscript{3}},
 \textbf{Xinhao Zhang\textsuperscript{3}},
\\
 \textbf{Kunpeng Liu\textsuperscript{3}},
 \textbf{Pengfei Wang\textsuperscript{1,2}\footnotemark[1]},
 \textbf{Yuanchun Zhou\textsuperscript{1,2}}
\\
\\
 \textsuperscript{1}Computer Network Information Center, CAS,
\\
 \textsuperscript{2}University of Chinese Academy of Sciences,
\\
 \textsuperscript{3}Portland State University
\\
 \texttt{wangzaitian23@mails.ucas.ac.cn}
}

\begin{document}
\maketitle
\setcounter{footnote}{0}
\renewcommand{\thefootnote}{\fnsymbol{footnote}}
\footnotetext[1]{Corresponding author}
\setcounter{footnote}{0}
\renewcommand{\thefootnote}{\arabic{footnote}}

\newcommand{\Methodname}{\textbf{DoAug}}
\newcommand{\Methodnamea}{\textbf{DoAug}}
\newcommand{\Methodnameb}{\textbf{DoAug}}
\newcommand{\Methodnamec}{\textbf{DoAug}}

\begin{abstract}

Data augmentation is an essential technique in natural language processing (NLP) for enriching training datasets by generating diverse samples. 
This process is crucial for improving the robustness and generalization capabilities of NLP models. 
However, a significant challenge remains: \textit{Insufficient Attention to Sample Distribution Diversity}. 
Most existing methods focus on increasing the sample numbers while neglecting the sample distribution diversity, which can lead to model overfitting. 
In response, we explore data augmentation's impact on dataset diversity and propose a \textbf{\underline{D}}iversity-\textbf{\underline{o}}riented data \textbf{\underline{Aug}}mentation framework (\textbf{DoAug}). 
Specifically, we utilize a diversity-oriented fine-tuning approach to train a large language model (LLM) as a diverse paraphraser, which is capable of augmenting textual datasets by generating diversified paraphrases. 
Then, we apply the LLM paraphraser to a selected coreset of highly informative samples and integrate the paraphrases with the original data to create a more diverse augmented dataset. 
Finally, we conduct extensive experiments on 12 real-world textual datasets. 
The results show that our fine-tuned LLM augmenter improves diversity while preserving label consistency, thereby enhancing the robustness and performance of downstream tasks. 
Specifically, it achieves an average performance gain of \(10.52\%\), surpassing the runner-up baseline with more than three percentage points. 


\end{abstract}

\section{Introduction}

\begin{figure}[t]
    \centering
    \includegraphics[width=0.9\linewidth]{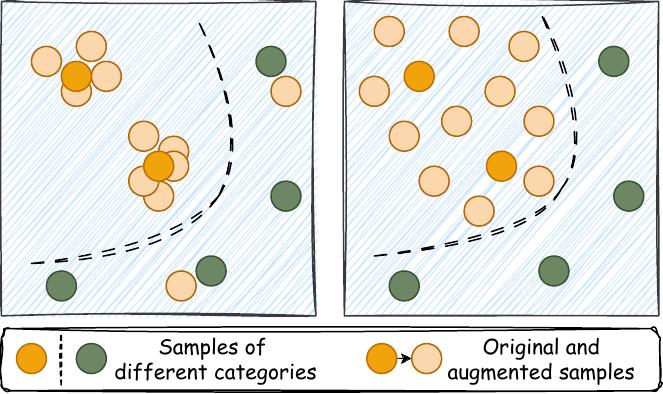}
    \caption{Conceptual comparison of \textbf{DoAug} (right) generating coherent and diverse samples against \underline{baselines} (left) generating noisy or repetitive samples. }
    \label{fig:doaug_abs}
\end{figure}

AI methods have demonstrated immense capabilities, often surpassing human abilities and traditional techniques across various natural language processing (NLP) tasks. 
This success largely hinges on the availability of high-quality datasets, which enable AI models to uncover intrinsic patterns and drive their effectiveness in real-world applications~\cite{xu2025scsiameseclu,zhang2025motif}. 
However, training on inferior datasets can significantly degrade model performance, particularly when applied to test data or real-world scenarios~\cite{wang2024comprehensive}. 
As AI technology advances, especially with large language models (LLMs), the demand for high-quality datasets has become more pronounced~\cite{wang2025sccompass}. 
To effectively train NLP models, a high-quality dataset should be 
(1) \textit{Large}: a sufficient number of samples is crucial to reflect the diversity and complexity of human language. 
Large datasets help prevent overfitting, ensuring the trained AI model generalizes well to unseen data. 
With more data points, the model can learn various patterns and relationships, which enhances its robustness and reliability; 
(2) \textit{Coherent}: the mapping between data and labels must be accurate and consistent. 
Coherent datasets ensure that each data point is correctly labeled, providing the model with reliable information for learning. 
Incoherent datasets, with mislabeled or inconsistent data, can confuse the model and degrade its performance. 
Consistency in labeling also aids in the reproducibility of task results and the interpretability of the model's predictions; 
(3) \textit{Diverse}: a diverse dataset ensures NLP models learn a broad spectrum of linguistic patterns, enhancing robustness across real-world conditions. 
This includes variations such as dialects, tones, formality, or domain-specific terms.
Exposure to such diversity helps models generalize better, avoiding over-reliance on narrow language subsets. 
Additionally, it improves adaptability to unexpected inputs while reducing biases tied to certain language styles.

Data augmentation is an efficient technique for increasing the number of training samples by modifying existing dataset samples~\cite{wang2024comprehensive}. 
It allows for the rapid generation of large-scale datasets without the need for additional data collection and has been successfully applied in various domains, including textual data. 
Data augmentation for textual data often changes the wording or reshapes the structure of a sentence. 
Many early and simple methods achieve this by randomly perturbing the textual samples at the word level~\cite{wei2019eda,karimi2021aeda}. 
While certain operations prove effective, some are prone to introducing noise that compromises label integrity (e.g. deleting a ``not'') or generating redundant samples, ultimately failing to improve dataset quality or promote diversity (Figure~\ref{fig:doaug_abs} left). 
Recent advances in LLMs have demonstrated unprecedented power in text understanding and generation~\cite{radford2018improving,chen2024large}. 
The capacity of these state-of-the-art (SOTA) generative language models establishes a new and promising paradigm of textual data augmentation~\cite{anaby2020not}. 
Generative models~\cite{brown2020language,ning2024fedgcs,dubey2024llama} enable large-scale acquisition of textual data while preserving the coherence of augmented datasets by generating texts with similar meanings to the original sentences~\cite{dai2025auggpt}.
However, most existing generative methods focus primarily on enlarging dataset size, with limited consideration of how the augmentation process affects diversity. 
Adequate attention to maintaining and enhancing dataset diversity is vital for developing AI-ready and high-quality datasets, making it a crucial challenge in designing textual data augmentation methods. 

Along this line, we propose a \textbf{\underline{D}}iversity-\textbf{\underline{o}}riented data \textbf{\underline{Aug}}mentation approach (\textbf{DoAug}) using an LLM to paraphrase sentences (Figure~\ref{fig:doaug_abs} right). 
The LLM is first fine-tuned on a paraphrase dataset and taught to rewrite sentences. 
By instructing the LLM to function as a paraphraser, we can use it to alter sentence expressions while preserving their essential meaning. 
In this way, we ensure the affinity between the original and augmented samples, minimizing the influence of data augmentation on dataset coherence. 
To enhance the dataset diversity through data augmentation, we further proposed a diversity-oriented fine-tuning method.  
We construct a preference dataset that chooses the more diverse paraphrases while rejecting repetitive ones. 
Then the LLM paraphraser is fine-tuned on the preference dataset with the DPO algorithm~\cite{rafailov2024direct} to encourage greater generation diversity. 
We also adopt a coreset selection method to focus on only the most important samples from the dataset to reduce the computational overhead and costs of running LLMs. 
Finally, we conduct extensive experiments on 12 textual benchmark datasets to verify the effectiveness of \Methodnamec. 
Experimental results show that our proposed method can remarkably enhance the diversity of the augmented dataset on 6 measurements while maintaining high affinity compared with the original dataset. 
We also investigate the implications of this increased diversity on model performance in downstream tasks and observe significant improvements in model performance when trained on the augmented datasets. 

In summary, the contributions are as follows: 
    
\textbf{(1)} We propose a data augmentation framework, \Methodname, that incorporates and explicitly encourages diversity, an important yet often neglected factor in high-quality datasets; 

\textbf{(2)} The framework trains and employs an LLM as a paraphraser to generate synthetic data with high affinity, ensuring the coherence of the augmented datasets; 

\textbf{(3)} We introduce a diversity-oriented fine-tuning method that trains the LLM augmenter on a preference dataset with the DPO algorithm to boost the generation diversity of the LLM;

\textbf{(4)} Extensive experiments conducted on 12 datasets demonstrate that \Methodnamec~significantly benefits learning performance by increasing dataset diversity while maintaining coherence. 

\section{Related Work}
\subsection{Textual Data Augmentation}
Textual data augmentation revolves around perturbing the wording and syntax of existing sentences to create more modified samples. 
Some early and simple methods propose to randomly replace, remove, insert, and swap characters or words at certain ratios in a sentence~\cite{belinkov2018synthetic,wei2019eda}. 
Some more sophisticated methods modify sentences by using alternative syntax~\cite{min2020syntactic}. 
Language models can in turn act as effective tools for textual data augmentation. 
For example, Back-translation~\cite{sennrich2016improving} first translates sentences from the source language (e.g. English) to an intermediary (e.g. Chinese) and then translates the intermediary sentence back to the source language. 
Substitute Word using BERT~\cite{kumar2020data} masks certain words in the original sentences and uses a BERT model to predict masked words. 
They utilize the subtle differences made by the translator or the unmasking process and perturb the wording or syntax while keeping the meanings untouched. 
The recent emergence of LLM has given birth to a series of new approaches~\cite{anaby2020not,cai2023resolving,ding2024data,wang2024survey}. 
AugGPT~\cite{dai2025auggpt}, for example, prompts the state-of-the-art ChatGPT model to rewrite sentences in the dataset and preserves dataset coherence after data augmentation. 
Self-LLMDA~\cite{li2024empowering} automatically generates and selects the most suitable instruction to prompt the LLM to generate augmented samples. 
However, these aforementioned methods neglect the impact on dataset diversity, failing to ensure the diversity trait of producing high-quality datasets. 
The effect of LLM augmentation diversity is discussed in \cite{cegin2023chatgpt,cegin2024effects}, where three types of prompt-based diversity incentives are proposed.

\subsection{Dataset Diversity Evaluation}
The evaluation of dataset diversity is increasingly popular as the size of available training data stunningly explodes, which makes it vital to maintain a minimized redundancy in the dataset to avoid repetitive training, saving the cost and time consumption and avoiding overfitting. 
Though its definition is not yet unified, many metrics are used across research. 
\cite{tevet2021evaluating} systematically studies the evaluation of text data diversity, which includes token-level metrics, embedding-level metrics, and human evaluations. 
\cite{lai2020diversity} proposes three dataset diversity metrics in the embedding space and investigates how these metrics change in different text datasets. 
\cite{yu2022can} proposes another three diversity metrics and discusses how improving dataset diversity helps enhance learning generalization, even when the total size of the dataset is reduced.
\cite{gontijo2020affinity} jointly investigates the role of data diversity and affinity in data augmentation, demonstrating that model performance benefits from improvements in both measures. 
Diversity has been considered in the design of several data augmentation methods~\cite{malandrakis2019controlled,liu2021divaug}, however, it has not yet been integrated with coherence-ensured and LLM-based data augmentation methods such as AugGPT~\cite{dai2025auggpt}. 

\begin{figure*}[htbp]
    \centering
    \includegraphics[width=0.9\linewidth]{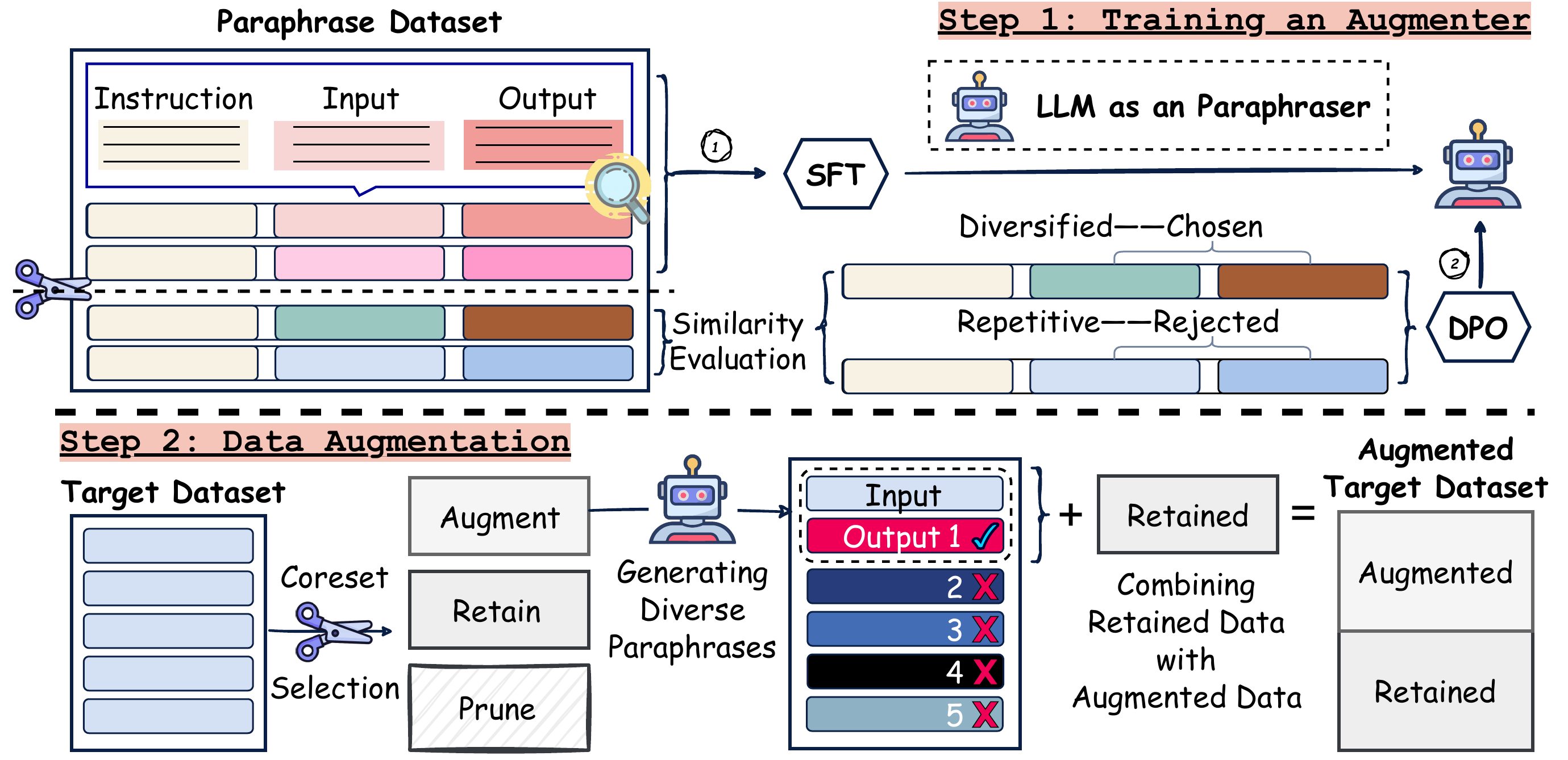}
    \caption{An overall framework of \Methodnamec.}
    \label{fig:framework}
\end{figure*}

\section{Methodology}
\subsection{Problem Formulation}

Given a parameterized data augmenter \(f_\theta\), the data augmentation process is expressed as \(f_\theta: \mathcal{S} = \{\mathbf{X}, \mathbf{t}\} \rightarrow \mathcal{\Tilde{S}} = \{\mathbf{\Tilde{X}}, \mathbf{\Tilde{t}}\} \), where \(\mathcal{S}\) is the original dataset composed of the feature vectors \(\mathbf{X}\) and target labels \(\mathbf{t}\), and \(\mathcal{\Tilde{S}}\) is the augmented dataset~\cite{wang2024comprehensive}. 
For a diversity metric \(D\), the diversity values of the original and the augmented datasets are \(D(\mathcal{S})\) and \(D(\mathcal{\Tilde{S}})\), respectively, and the diversity gain of that augmentation is defined as \(\Delta D(\mathcal{S}; \theta) = D(\mathcal{\Tilde{S}}) - D(\mathcal{S})\).

\Methodnamea~aims to optimize the parameter \(\theta^{\ast}\) for the data augmenter to maximize the diversity gain after augmentation:
\begin{equation}
\theta^{\ast} = \mathop{\arg \max}_{\theta} \mathbbm{E} \Delta D(\mathcal{S}; \theta)
\end{equation}
When training models on the original and augmented datasets, their respective performances are evaluated by a performance metric \(P\), resulting \(P(\mathcal{S})\) for the original dataset and \(P(\mathcal{\Tilde{S}})\) for the augmented dataset. 
The performance gain is defined as \(\Delta P(\mathcal{S}; \theta) = P(\mathcal{\Tilde{S}}) - P(\mathcal{S})\).  
Moreover, by optimizing and employing the augmenter with maximum diversity gain, \Methodnamec~expects to achieve maximum performance gain under fixed conditions: 
\begin{equation}
\Delta P(\mathcal{S}; \theta^{\ast}) = \mathop{\max}_{\theta} \Delta P(\mathcal{S}; \theta)
\end{equation}

\subsection{Framework Overview}

\Methodnamea~trains and employs an LLM capable of generating diverse paraphrases for data augmentation to enlarge the size of textual datasets, maintain coherence, and enhance diversity\footnote{Code is available at: \url{https://github.com/CNICDS/DoAug}}. 
As Figure~\ref{fig:framework} shows, the framework is organized as follows:
    
\noindent$\bullet$ An LLM is first trained on a paraphrase dataset through supervised instruction fine-tuning, enabling it to function as a paraphraser that rewrites sentences while preserving the original semantics; 

\noindent$\bullet$ The LLM paraphraser is then trained on a constructed preference dataset with the DPO (Direct Preference Optimization) algorithm that encourages diverse generation samples; 

\noindent$\bullet$ For a given textual dataset, a coreset of samples is selected based on their importance, serving as the source sentences for paraphrase augmentation; 

\noindent$\bullet$ Generated paraphrases are ranked and sampled according to their diversity, with the most diverse paraphrases integrated into the augmented dataset with original coreset samples. 

\subsection{Diverse Paraphraser Fine-tuning}

The proposed data augmentation method is constructed upon a general-purpose LLM. 
Pre-trained on a vast amount of corpus, LLMs now boast human-level understanding and generation abilities~\cite{ouyang2022training,dubey2024llama}. 
We leverage these abilities of an LLM and use it as a tool for our data augmentation process. 
The LLM is fine-tuned as a paraphraser that can rewrite sentences with alternative expressions while maintaining the original semantics. 
The LLM paraphraser is further fine-tuned to produce more diverse generation results and cover more linguistic alternatives. 
On the one hand, the change in sentence expressions can introduce diversity to the dataset. 
On the other hand, since the LLM is capable of capturing the semantics of the sentence and is prompted only to paraphrase the sentence, the coherence of the augmented data-label mapping is preserved. 

\subsubsection{LLM Paraphraser Training with PEFT} 
To fully leverage the understanding and generation abilities of the LLM, we use supervised fine-tuning (SFT) to train it to follow instructions to paraphrase existing sentences. 
Since the SFT phase of LLM training is heavily computation-consuming, we use the Parameter-Efficient Fine-Tuning (PEFT) technique to reduce the size of trainable parameters updated in the back-propagation pass to save the computation cost. 
Specifically, we adopt the Low-Rank Adaptation (LoRA) approach~\cite{hu2021lora}. 
Given a pre-trained LLM with weights \(W_0 \in  \mathbb{R}^{d \times k}\), LoRA represents its update \(\Delta W\) with \(BA\), where \(B \in \mathbb{R}^{d \times r}, A \in \mathbb{R}^{r \times k}\), and the rank \(r \ll min(d,k)\). 
During training, \(W_0\) is frozen and excluded from the gradient update, while \(A\) and \(B\) are updated instead. 
After training, \(W_0\) and \(\Delta W = BA\) are multiplied with the same input, and their outputs are summed, as in:
\begin{equation}
    h = W_0 x + \Delta Wx = W_0 x + B A x
\end{equation}
In the SFT phase, we sample a subset \(\mathcal{D}_{\text{SFT}}\) from ChatGPT Paraphrases dataset\footnote{https://huggingface.co/datasets/humarin/chatgpt-paraphrases} to train the LLM. 

\subsubsection{LLM Generation Diversity Enhancement with DPO}

To align LLM generations to human preferences, recent LLMs adopt RLHF in their training process, which involves the PPO algorithm and a reward model~\cite{ouyang2022training,zhang2024tfwt,zhang2024prototypical}. 
However, fitting a reward model brings extra computation costs, and the gap between its prediction and actual human preference also poses threats to the effect of PPO. 
As an alternative, the DPO algorithm directly optimizes the LLMs' generation policy without training an additional reward model~\cite{rafailov2024direct}. 

\paragraph{Preference Dataset Construction.}
To construct a dataset \(\mathcal{D}_{\text{DPO}}\) for DPO training, we sample another subset from the original paraphrase dataset. Each original sentence corresponds to 5 paraphrases, and we embed the original sentence \(x\) and paraphrased sentences \([y_1, ..., y_5]\) and calculate the Euclidean distances in the embedding space \(\mathcal{E}\):
\begin{equation}
    dist(y_i, x) = \sqrt{(e_{y_i} - e_x)^2},
\label{eq:dist}
\end{equation}
where \(e_x = \mathcal{E}(x)\). The paraphrase with maximum distances is considered the most diverse among possible generation results and used as the ``chosen'' (preferred) output. In contrast, the most similar is taken as the least varied generation and used as the ``rejected'' (dispreferred) one. This preference construction process is formulated in Eq.~\ref{eq:pref}: 
\begin{equation}
    \begin{aligned}
    &y_w= \arg \max_{y_i} dist(y_i,x) \\
    \vspace{-2mm}
    &y_l= \arg \min_{y_i} dist(y_i,x),
    \end{aligned}
\label{eq:pref}
\end{equation}
where \(y_w\) is the preferred paraphrase out of the pair \((y_w, y_l)\).

\paragraph{Training Objective.}
The goal of DPO training is to maximize the probability of generating the preferred output and minimize the probability of generating the dispreferred output. Unlike the PPO algorithm which requires a reward model and a reinforcement learning phase, DPO derives its objective by solving the optimal solution of PPO's optimization problem, as shown in Eq.~\ref{eq:dpo}:
\begin{multline}
    \mathcal{L}_\text{DPO} = 
    - \mathbb{E}_{(x, y_w, y_l) \sim \mathcal{D}_{\text{DPO}}}
     \bigg[ \log \sigma \bigg(
    \\
    \beta \log \frac{\pi_{\theta}(y_w | x)} 
    {\pi_{\text{ref}}(y_w | x)} 
    \left. \left. - \beta \log \frac{\pi_{\theta}(y_l | x)} 
    {\pi_{\text{ref}}(y_l | x)} 
     \right) \right],
\label{eq:dpo}
\end{multline}
where \(\pi_\theta\) denotes the generation probability of the current LLM, \(\pi_{\text{ref}}\) denotes that of the SFT model, and hyper-parameter \(\beta\) controls the deviation from the SFT model. 

\subsubsection{Diversity-oriented Sampling}

For each input sentence, we use beam search to generate \(K\) sequences from the LLM's output logits. 
We then rank these sequences based on their distances from the original sentences according to Eq.~\ref{eq:dist}. 
Only the most distant sentences are retained to reduce redundancy and prevent overfitting.

\begin{algorithm}[H]
    \caption{Diversity-oriented and Coreset-focused Data Augmentation}
    \label{algo}
    \begin{algorithmic}[1]
    \REQUIRE An LLM \(f_{\theta}\), a paraphrase dataset \(\mathcal{D}_{\text{SFT}}\), a preference dataset \(\mathcal{D}_{\text{DPO}}\), and a target textual dataset \(\mathcal{S}\)
    \STATE Train the original LLM \(f_{\theta}\) on the paraphrase dataset \(\mathcal{D}_{\text{SFT}}\) with SFT
    \STATE Fine-tune the LLM on the preference dataset \(\mathcal{D}_{\text{DPO}}\) by optimizing the loss function in Eq.~\ref{eq:dpo}
    \STATE Initialize empty \(\mathcal{S}^\prime\)
    \FORALL{\((x,t) \in \mathcal{S}\) }
        \STATE Calculate importance score \(s\) for \(x\)
        \STATE \(\mathcal{S}^\prime\).\texttt{append}(\((x,t,s)\))
    \ENDFOR
    \STATE Rank samples in \(\mathcal{S}^\prime\) according to the score \(s\)
    \STATE Split \(\mathcal{S}^\prime\) with ratio \(r_\text{augment} : r_\text{retain} : r_\text{prune}\) to \(\mathcal{S}_{\text{augment}}\), \(\mathcal{S}_{\text{retain}}\), and \(\mathcal{S}_{\text{prune}}\)
    \STATE Initialize empty \(\mathcal{\Tilde{S}}\)
    \FORALL{\((x,t,s) \in \mathcal{S}_{\text{augment}}\)}
        \STATE \(y = f_{\theta} (x)\)
        \STATE \(\mathcal{\Tilde{S}}\).\texttt{append}(\((x,t)\))
        \STATE \(\mathcal{\Tilde{S}}\).\texttt{append}(\((y,t)\))
    \ENDFOR
    \STATE \(\mathcal{\Tilde{S}} = \mathcal{\Tilde{S}} \cup \mathcal{S}_{\text{retain}}\)
    \RETURN Augmented dataset \(\mathcal{\Tilde{S}}\)
    \end{algorithmic}
\end{algorithm}

\subsection{Selective Coreset Data Augmentation}

Given the time consumption and computation cost of LLM-based data augmentation methods, it is non-trivial to recognize the most important samples and constrain the target for data augmentation to these samples. 
First, we train the downstream task model on the dataset and collect training dynamics and post-training post-training metrics. 
Then we calculate the EL2N~\cite{paul2021deep}, entropy~\cite{coleman2020selection}, variance~\cite{swayamdipta2020dataset}, and AUM~\cite{pleiss2020identifying} score to evaluate sample importance. 
We use score monotonic selection and coverage-centric selection (CCS)~\cite{zheng2023coverage} to derive the coresets.
\Methodnamea~performs a hierarchical coreset selection to prune some low-importance samples, retain the middle-importance samples, and augment the high-importance samples. 
Only samples of high importance are used as the seeds for data augmentation. 
The original sentences in the high-importance coreset and their paraphrases are combined with the middle-importance samples, composting the final results of our data augmentation process, as presented in Algorithm~\ref{algo}.

\section{Experiments}

\subsection{Experiment Settings}

\subsubsection{Evaluation Criterion}
We evaluate \textit{\textbf{diversity}} and \textit{\textbf{affinity}} for data distribution while measuring \textit{\textbf{performance}} on downstream tasks for effectiveness. For all these measurements, the higher score indicates better results. 

\noindent \textit{\textbf{Diversity.}}
To comprehensively evaluate \Methodname's effect on dataset diversity, we adopt several measurements to assess the augmented dataset's diversity from both latent and lexical aspects: 

\(\bullet\) \textit{Distance} and \textit{Dispersion} assess datasets' latent diversity at the sample level by calculating pair-wise Euclidean distance and Cosine similarity on the embedding space. 

\(\bullet\) \textit{Isocontour Radius} and \textit{Homogeneity} assess datasets' latent diversity at the dataset level by considering the coverage and uniformity of all sample embeddings. 

\(\bullet\) \textit{Vocabulary Size} and \textit{Unique 3-grams} assess datasets' lexical diversity by counting how many different words are used throughout the datasets. 

\noindent\textit{\textbf{Affinity.}} The affinity score reflects the coherence of an augmented dataset and is embodied by embedding deviation. 

\noindent\textit{\textbf{Performance}} on downstream task. 
Following the practice in existing research on textual data augmentation, we train a \(\text{BERT}_{\text{base}}\) model~\cite{kenton2019bert} with a classification head on the original and augmented datasets to evaluate the effect of our proposed data augmentation approach. 
We report the prediction accuracy scores on each dataset to measure downstream task performance.

\subsubsection{Datasets} 
We conduct extensive experiments on 12 NLP datasets to verify the effectiveness of \Methodnamec. 
Our selection of datasets covers a wide range of text classification tasks, including entailment annotation (ANLI, MNLI, and RTE), sentiment analysis (MPQA, SST-2, and Yelp), chemical-protein relationship (ChemProt), acceptability judgment (CoLA), semantically equivalence (MRPC), sentence role (RCT), subjectivity analysis (SUBJ), and Disease judgment (Symptoms)~\cite{pang2004sentimental,wiebe2005annotating,zhang2015character,kringelum2016chemprot,dernoncourt2017pubmed,wang2019glue,nie2020adversarial,dai2025auggpt}.
More details of these datasets are specified in Appendix~\ref{app:dataset}. 
Following the settings in~\cite{yoo2021gpt3mix}, we sample a subset (1.2K samples) from the full dataset to unify the evaluation settings, enable a fair comparison between methods, and simulate a low-resource condition where data augmentation is of significant necessity. 

\subsubsection{Baseline Methods}
We compare \Methodnamec~with twelve representative data augmentation methods. 
(1) OCR and (2) Keyboard perform common OCR or typing errors at the character level~\cite{li2024empowering}. 
(3) EDA randomly inserts, deletes, replaces, or swaps words in the sentences~\cite{wei2019eda}. 
(4) AEDA randomly inserts punctuations in the sentences~\cite{karimi2021aeda}. 
(5) Back-translation (BT) involves translating the source sentences to an intermediary language~\cite{sennrich2016improving}. 
(6) Unmask randomly replaces words with \texttt{[MASK]} and predicts the masked words with the BERT model~\cite{kumar2020data}. 
(7) AugGPT directly prompts ChatGPT (replaced with Llama3.1-8B-Instruct to save cost) for paraphrases without further fine-tuning~\cite{dai2025auggpt}.
(8) Grammar and (9) Spelling are two exemplar methods selected by Self-LLMDA, which prompt the LLM to simulate common grammatical variation or spelling errors made by humans~\cite{li2024empowering}.
(10) Chain, (11) Hint, and (12) Taboo generate paraphrases with three different diversity incentives~\cite{cegin2024effects}. 
Augmentation examples of these methods are presented and discussed in Appendix~\ref{app:example} Table~\ref{tab:examples}. 

\subsubsection{Key Implementation Information}

We use Llama-3.2-1B-Instruct with BF16 quantization as the LLM paraphraser. 
The LLM's training dataset \(\mathcal{D}_{\text{SFT}}\) and \(\mathcal{D}_{\text{DPO}}\) contain 100,000 sentence pairs and 50,000 preference pairs, respectively. 
In the coreset selection step, the coreset ratio \(r_\text{augment} : r_\text{retain} : r_\text{prune}\) is \(1:1:1\). We explain how we derive this ratio in Appendix~\ref{app:ratio}.
For each original sentence, we generate \(K=5\) paraphrases and sample the most diversified output. 
A more detailed implementation specification is given in Appendix~\ref{app:implemtation}. 

\begin{table*}[thbp]
\resizebox{\textwidth}{!}{
    \centering
    \tiny
    \setlength{\tabcolsep}{1.2pt} 
    \renewcommand{\arraystretch}{1.0} 
\begin{tabular}{ccccccccccccccc}
    \toprule
    ~                      & ANLI               & ChemProt           & CoLA               & MNLI               & MPQA               & MRPC               & RCT                & RTE                & SST-2              & SUBJ               & Symptoms           & Yelp               & Avg. Gain \\
    \midrule
    Original       & \(35.75_{1.68}\)   & \(58.33_{4.67}\)   & \(74.56_{1.11}\)   & \(42.81_{2.33}\)   & \(89.17_{0.47}\)   & \(76.50_{3.09}\)   & \(71.62_{2.49}\)   & \(53.61_{2.45}\)   & \(86.97_{1.00}\)   & \(\underline{95.75}_{0.23}\) & \(74.06_{9.86}\)   & \(51.48_{6.81}\)   & -             \\
    OCR            & \(34.43_{1.50}\)   & \(63.84_{3.14}\)   & \(73.12_{1.82}\)   & \(53.94_{8.10}\)   & \(88.83_{0.48}\)   & \(75.49_{4.48}\)   & \(\underline{79.66}_{1.02}\)   & \(56.06_{3.92}\)   & \(86.91_{0.74}\)   & \(95.25_{0.25}\)   & \(86.12_{2.72}\)   & \(55.46_{1.04}\)   & \(4.73\%\)    \\
    Keyboard       & \(34.99_{1.53}\)   & \(65.78_{2.28}\)   & \(72.69_{1.19}\)   & \(52.88_{4.44}\)   & \(88.98_{0.43}\)   & \(77.99_{2.74}\)   & \(79.42_{0.72}\)   & \(57.58_{4.23}\)   & \(86.12_{0.66}\)   & \(95.25_{0.37}\)   & \(87.24_{1.09}\)   & \(55.44_{0.62}\)   & \(5.40\%\)    \\
    EDA            & \(34.91_{2.52}\)   & \(64.24_{2.45}\)   & \(73.07_{0.95}\)   & \(55.90_{3.21}\)   & \(89.12_{0.60}\)   & \(79.44_{2.11}\)   & \(77.17_{1.28}\)   & \(55.96_{3.85}\)   & \(87.48_{1.07}\)   & \(95.63_{0.27}\)   & \(89.11_{1.45}\)   & \(55.15_{1.34}\)   & \(6.68\%\)    \\
    AEDA           & \(35.31_{1.94}\)& \(64.87_{2.14}\)& \(72.29_{1.09}\)& \(57.23_{0.96}\)& \(89.15_{0.27}\)& \(79.41_{1.41}\)& \(78.48_{0.34}\)& \(54.51_{2.40}\)& \(86.03_{1.43}\)& \(95.24_{0.31}\)& \(89.66_{0.88}\)& \(54.52_{0.31}\) & \(6.75\%\) \\
    BT             & \(34.39_{0.78}\)   & \(67.72_{2.83}\)   & \(70.42_{2.03}\)   & \({56.40}_{3.19}\) & \(\underline{89.73}_{0.38}\) & \(78.55_{2.62}\)   & \(76.36_{1.09}\)   & \(53.52_{3.67}\)   & \(85.89_{1.42}\)   & \(95.16_{0.39}\)   & \(89.43_{0.59}\)   & \({55.56}_{0.85}\) & \(6.26\%\)    \\
    Unmask         & \(36.04_{1.19}\)   & \(69.36_{1.85}\)   & \(74.11_{1.10}\)   & \(54.60_{5.77}\)   & \(88.87_{0.45}\)   & \(\underline{80.15}_{1.27}\)   & \(79.01_{0.87}\)   & \(55.85_{3.97}\)   & \(87.08_{1.00}\)   & \(95.22_{0.21}\)   & \(89.92_{0.55}\)   & \(55.10_{0.69}\)   & \(6.59\%\)    \\
    AugGPT         & \(\underline{36.43}_{1.04}\) & \(65.73_{2.80}\)   & \(\underline{75.17}_{1.35}\) & \(53.77_{2.61}\)   & \(89.67_{0.34}\)   & \(75.25_{2.12}\)   & \({78.90}_{0.70}\) & \(54.87_{2.51}\)   & \(\underline{87.63}_{0.60}\) & \(95.44_{0.31}\)   & \(79.25_{3.33}\)   & \(55.47_{0.64}\)   & \(5.64\%\)    \\
    Grammar        & \(35.39_{1.55}\)   & \(68.64_{2.47}\)   & \(72.16_{1.45}\)   & \(56.53_{1.30}\)   & \(89.46_{0.42}\)   & \(78.90_{1.83}\)   & \(77.35_{0.68}\)   & \(54.40_{3.58}\)   & \(86.62_{1.56}\)   & \(94.98_{0.25}\)   & \(\underline{89.98}_{0.43}\)   & \(55.27_{0.82}\)   & \(5.88\%\)    \\
    Spelling       & \(35.98_{1.35}\)   & \(68.69_{2.06}\)   & \(72.09_{1.66}\)   & \(56.76_{1.04}\)   & \(88.96_{0.52}\)   & \(79.12_{1.77}\)   & \(78.95_{0.66}\)   & \(57.40_{3.09}\)   & \(86.42_{0.89}\)   & \(95.20_{0.42}\)   & \(89.48_{0.92}\)   & \(54.76_{0.99}\)   & \(6.49\%\)    \\
    Chain          & \(35.25_{1.82}\)   & \(68.75_{2.66}\)   & \(72.32_{0.86}\)   & \(56.23_{1.64}\)   & \(89.10_{0.41}\)   & \(75.88_{2.85}\)   & \(79.31_{0.54}\)   & \(54.98_{5.42}\)   & \(86.79_{0.82}\)   & \(95.14_{0.36}\)   & \(89.24_{0.55}\)   & \(55.50_{0.66}\)   & \(5.77\%\)    \\
    Hint           & \(36.19_{1.24}\)   & \(68.67_{1.96}\)   & \(72.25_{1.47}\)   & \(56.69_{1.47}\)   & \(89.13_{0.67}\)   & \(78.75_{2.07}\)   & \(78.44_{0.41}\)   & \(55.78_{2.98}\)   & \(86.80_{0.74}\)   & \(95.00_{0.31}\)   & \(89.58_{1.35}\)   & \(55.88_{0.32}\)   & \(6.43\%\)    \\
    Taboo          & \(35.83_{1.75}\)   & \(\underline{69.66}_{1.79}\)   & \(72.90_{1.21}\)   & \(\underline{57.26}_{0.93}\)   & \(89.34_{0.29}\)   & \(76.74_{2.30}\)   & \(78.48_{0.41}\)   & \(\mathbf{58.01}_{3.41}\)   & \(86.74_{1.43}\)   & \(95.12_{0.34}\)   & \(89.40_{0.64}\)   & \(\underline{56.30}_{0.71}\)   & \(\underline{6.76\%}\)    \\
    \midrule
    \Methodnameb~ & \(\mathbf{38.46}_{2.51}\) & \(\mathbf{70.22}_{1.76}\) & \(\mathbf{75.62}_{0.51}\) & \(\mathbf{59.76}_{1.02}\) & \(\mathbf{89.78}_{0.29}\) & \(\mathbf{80.97}_{3.45}\) & \(\mathbf{80.10}_{0.63}\) & \(\underline{56.05}_{2.72}\) & \(\mathbf{88.64}_{0.80}\) & \(\mathbf{95.80}_{0.19}\) & \(\mathbf{90.74}_{1.85}\) & \(\mathbf{56.57}_{0.49}\) & \(\mathbf{10.52\%}\) \\

    \bottomrule
\end{tabular}
}
    \caption{Prediction accuracy of models trained on augmented datasets. The best results are highlighted with the \textbf{bold} font, and runner-ups are \underline{underlined}. We report the mean performance and standard deviation and the results are averaged on ten random seeds. }
    \label{tab:results}
\end{table*}

\begin{table}[tbp]
    \centering
    \tiny
    \setlength{\tabcolsep}{1pt} 
    \begin{tabular}{lccccccc}
        \toprule
              &   {Distance }   &   {Dispersion }    &   {Radius }      &   {Homogeneity }     &   {Vocabulary } 
              & {3-grams } & {Average } \\

\midrule

Original & 0.00 & 0.00 & 0.78 & 0.74 & 0.00 & 0.00 & 0.25 \\
OCR & 0.01 & 0.05 & 0.62 & 0.83 & 0.05 & 0.11 & 0.28 \\
Keyboard & 0.00 & 0.05 & 0.55 & 0.83 & 0.08 & 0.17 & 0.28 \\
EDA & 0.27 & 0.44 & 0.00 & 0.86 & 0.19 & 0.46 & 0.37 \\
AEDA & 0.02 & 0.09 & 0.17 & 0.95 & 0.00 & 0.14 & 0.23 \\
BT & 0.38 & 0.42 & 0.70 & 0.54 & 0.36 & 0.59 & 0.50 \\
Unmask & 0.09 & 0.11 & 0.69 & 0.83 & 0.05 & 0.25 & 0.33 \\
AugGPT & 0.23 & 0.19 & 0.92 & 0.47 & 0.24 & 0.31 & 0.39 \\
Grammar & \underline{0.64} & \underline{0.62} & \textbf{1.00} & 0.00 & 0.13 & 0.54 & 0.49 \\
Spelling & 0.14 & 0.17 & 0.47 & 0.83 & \underline{0.49} & 0.37 & 0.41 \\
Chain & 0.27 & 0.19 & \underline{0.98} & 0.95 & 0.48 & 0.67 & 0.59 \\
Hint & 0.56 & 0.51 & \underline{0.98} & 0.86 & 0.45 & \underline{0.68} & \underline{0.67} \\
Taboo & 0.26 & 0.18 & 0.94 & \textbf{1.00} & 0.35 & 0.59 & 0.55 \\

\midrule
\textbf{DoAug} & \textbf{1.00} & \textbf{1.00} & 0.87 & \underline{0.98} & \textbf{1.00} & \textbf{1.00} & \textbf{0.98} \\

        \bottomrule 
    \end{tabular}
    \caption{6 diversity metrics averaged on 12 datasets and the average score, normalized to \([0,1]\).}
    
    \label{tab:diversity_norm}
\end{table}

\subsection{Overall Results}

\begin{figure}[htbp]
    \centering
    \includegraphics[width=0.95\linewidth]{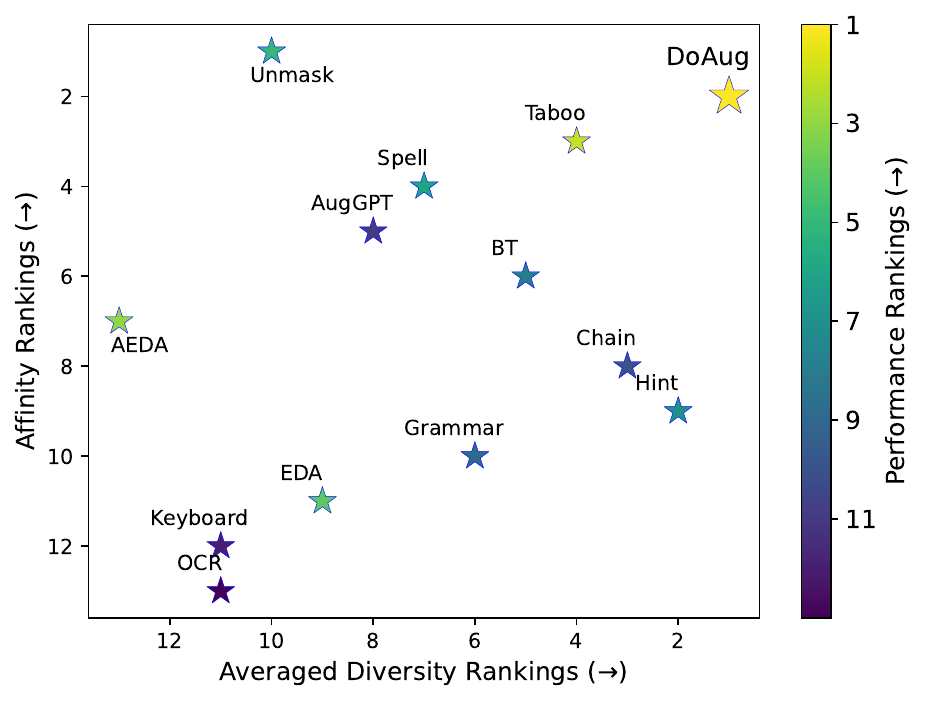}
    \caption{Diversity, affinity, and performance achieved by \Methodnamec~and baseline methods. Results are averaged on 12 datasets and the diversity rankings are further averaged on 6 metrics in this diagram. A smaller number for the rankings indicates better results. }
    \label{fig:overall}
\end{figure}

To verify the effectiveness of \Methodnamec, we evaluate downstream task accuracy alongside the diversity and affinity of the augmented dataset. 
We report the rankings of performance, diversity, and affinity averaged on 12 datasets achieved by \Methodnamec~and 12 baseline methods in Figure~\ref{fig:overall}. 
From these results, we have the following observations: 
\textbf{(1)} \textbf{\Methodnamea~achieves the highest performance on downstream tasks} compared to other SOTA data augmentation methods, as indicated by the color bar in Figure~\ref{fig:overall}. 
This demonstrates the high quality and superior adaptability of the datasets generated by our proposed method in real-world applications. 
\textbf{(2)} \textbf{\Methodnamea~achieves the highest diversity score and outperforms all other baseline methods.} This implies that \Methodnamea~effectively improves dataset diversity. 
\textbf{(3)} \textbf{\Methodnamea~achieves a considerably high position on the affinity rankings}, indicating that the sample semantics are preserved to the greatest extent possible. 
In sum, \Methodnamea~achieves the top position in the combined dataset diversity and affinity rankings. 
Additionally, it achieves the best downstream task performance, indicated by the lightest yellow. 

\subsection{Performance, Diversity, and Affinity}

\subsubsection{Performance Gains}

The full results for BERT classification performance on original and augmented datasets are presented in Table~\ref{tab:results}. The results show that \Methodnamec~surpasses all baseline methods on average. Specifically, it outperforms the baseline methods on 11 out of 12 datasets except on RTE. \Methodnamea~achieves performance gain of \(10.52\%\) on average, surpassing the runner-up method with an advantage of 3.76 percentage points. 

\subsubsection{Diversity Gain}

We demonstrate the diversity gains in terms of all 6 diversity metrics achieved by \Methodnamec~and baseline methods in Figure~\ref{tab:diversity_norm}. 
For each metric, the scores are normalized to \([0,1]\), and we also include the original scores in Appendix~\ref{app:diversity}.
\Methodnamec~ranks the top on the chart with an average score of 0.98. Specifically, it achieves the best for the Distance, Dispersion, Vocabulary Size, and Unique 3-grams metrics. It is also competitive in terms of Isocontour Radius and Homogeneity. The three baselines with diversity incentives, namely Chain, Hint, and Taboo, also achieve reasonably good diversity gain, in line with the results of \cite{cegin2024effects}.

\begin{figure}[t]
    \centering
    \includegraphics[width=0.98\linewidth]{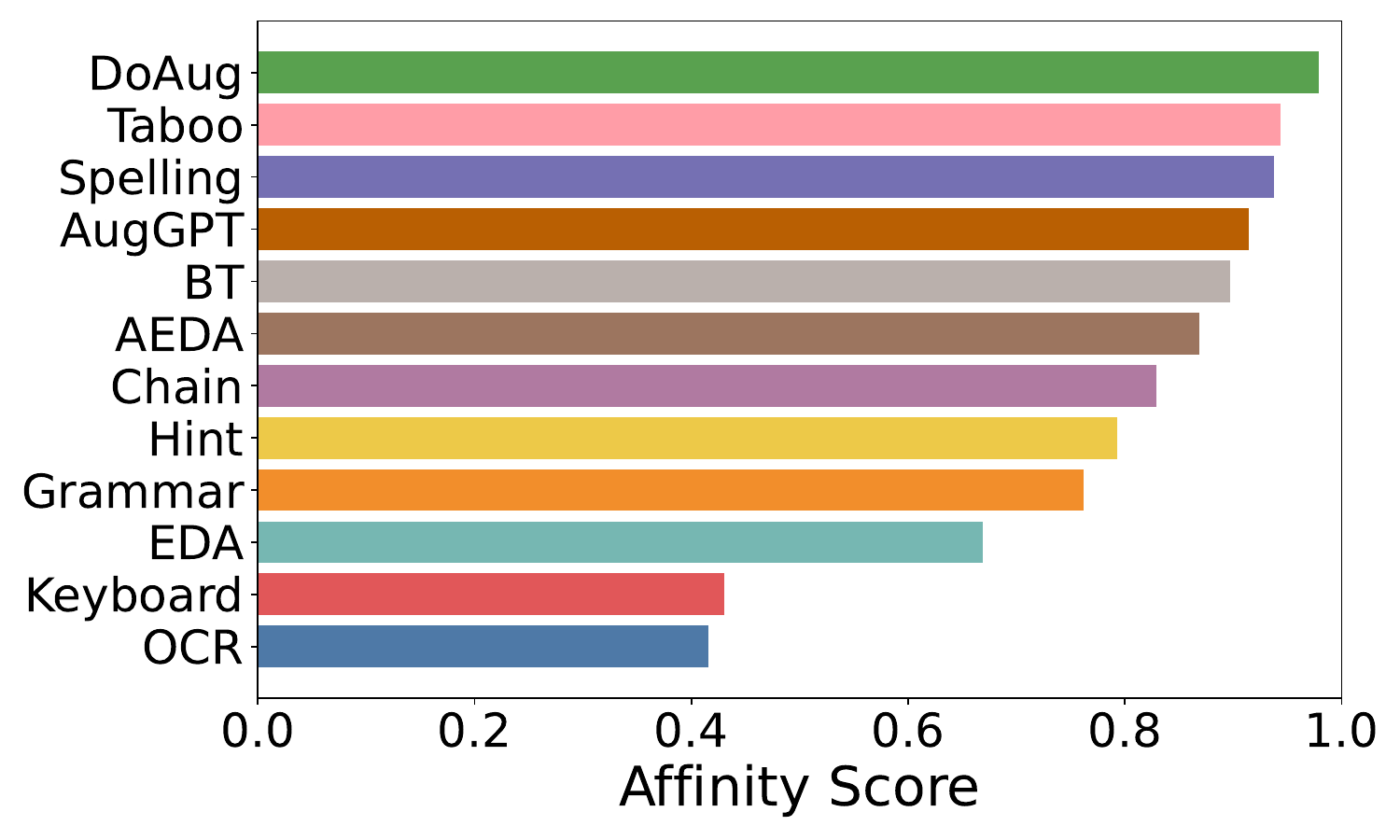}
    \captionof{figure}{Affinity scores of \Methodnamec~and 10 baseline methods. The scores are averaged on 12 datasets.}
    \label{fig:affinity}
\end{figure}

\subsubsection{Affinity and Paraphrase Validity}

We present the affinity of \Methodnamec~and baseline methods in Figure~\ref{fig:affinity}, where \Methodnamea~outperforms other methods except Unmask, whose affinity score is 1.95. The Unmask method generates augmentation by replacing randomly selected words with ``\texttt{[MASK]}'' and predicts the masked words with the BERT model. Since the augmented samples are recovered from the BERT embeddings of the corrupted original samples and we use the BERT embeddings to calculate affinity, it is reasonable to yield extremely high affinity scores. 
Following \cite{cegin2023chatgpt,cegin2024effects}, we also investigate paraphrase validity at the sample level. 
We perform a human evaluation on 200 samples per dataset to check if the paraphrases are semantically similar to the original samples and adhere to the original labels. Results show \(95\%\) paraphrases are valid. 
We also prompt the DeepSeek-V3 model, a very strong and knowledgeable LLM that is good at understanding users' intent and evaluating the task scenario for an LLM-based evaluation. Results show that \(97\%\) paraphrases are valid, suggesting that \textbf{DoAug} introduces negligible noises to the dataset. 

\subsection{Ablation Studies}

\begin{figure}[t]
    \centering
    \includegraphics[width=0.85\linewidth]{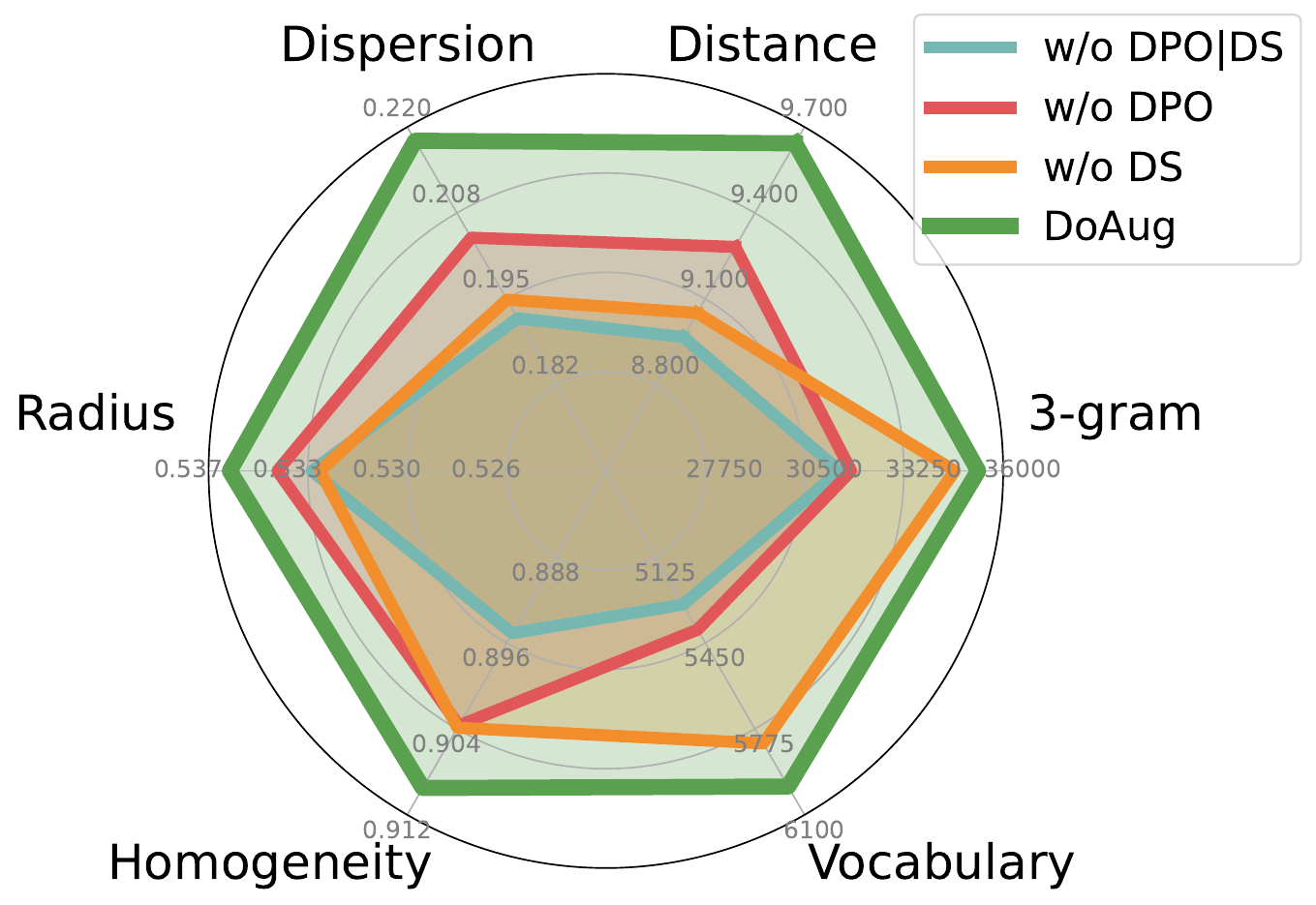}
    \caption{Ablation study on diversity gains}
    \label{fig:div_ablt}
\end{figure}

\begin{figure}[t]
    \centering
    \subfloat[Accuracy]{
        \includegraphics[width=0.30\linewidth]{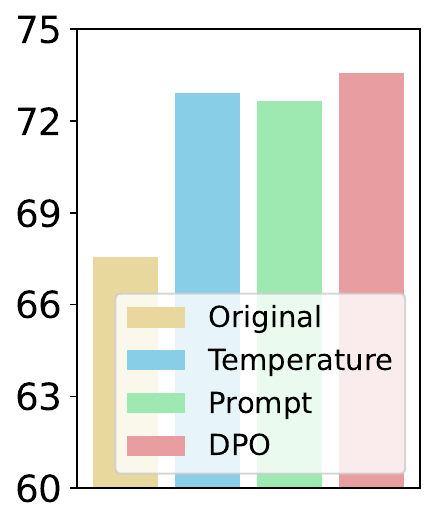}
        \label{subfig:promptemp_perf}
    }
    \subfloat[Distance]{
        \includegraphics[width=0.31\linewidth]{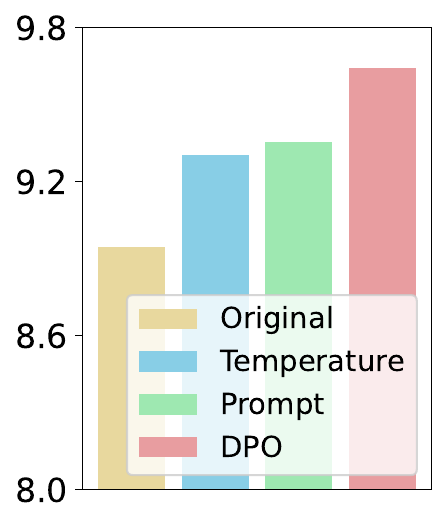}
        \label{subfig:promptemp_dist}
    }
    \subfloat[3-grams]{
        \includegraphics[width=0.35\linewidth]{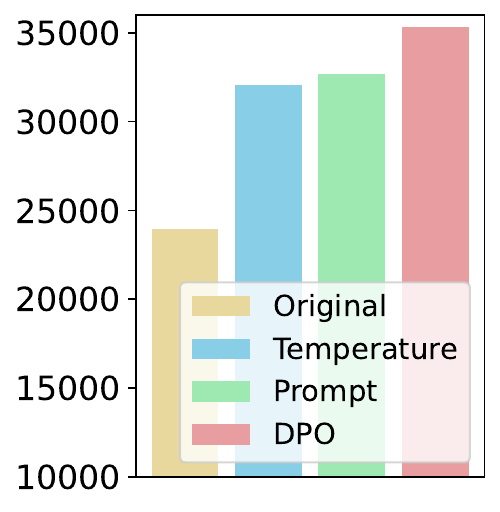}
        \label{subfig:promptemp_3gram}
    }
    \caption{Replacement study on DPO training.}
    \label{fig:promptemp}
\end{figure}

To verify the effectiveness of our proposed method, we conduct ablation studies as shown in Table~\ref{tab:ablation}, where w/o Coreset refers to applying augmentation on a random subset of the dataset without deriving a coreset of importance samples, w/o Selective refers to augmenting samples in both \(\mathcal{S}_\text{retain}\) and \(\mathcal{S}_\text{augment}\) instead of only augmenting the latter, w/o Aug refers to using the coreset directly for training without data augmentation, w/o DPO refers to using the LLM paraphraser from the SFT stage for data augmentation, w/o DS refers to removing the diversity-based sampling module, and w/o DPO|DS removes both steps. 
From these results, we find out that all components in the method framework make remarkable contributions to the final performance gains. 
We also notice that focusing augmentation on the selected coreset is the most important factor of performance (with the lowest average scores and no runner-up results). 
As Figure~\ref{fig:div_ablt} shows, we also study the effect of diversity-oriented fine-tuning and diversity-based sampling on the dataset diversity, demonstrating that all proposed components are effective. 
Concretely, we can observe that diversity sampling has more influence on sample-level latent diversity, while DPO training has more influence on lexical diversity. 
Further, we investigate whether the DPO training can be replaced by cost-saving approaches, such as sampling from a higher temperature and using prompts with diversity incentives. 
Results in Figure~\ref{fig:promptemp} indicate that replacing DPO training with sampling from a higher temperature or using prompts with diversity incentives achieves inferior results, failing to compete with the DPO version, suggesting model fine-tuning is necessary for diversity gain and performance improvement. Full results are in Appendix~\ref{app:promptemp_app}.

\begin{table}[t]
\resizebox{\linewidth}{!}{
    \centering
    \tiny
    \setlength{\tabcolsep}{0.6pt} 
    \renewcommand{\arraystretch}{1.2} 
    \begin{tabular}{lcccccccccccc}
    \toprule
                      & ANLI  & Ch.Pr. & CoLA  & MNLI  & MPQA  & MRPC  & RCT   & RTE   & SST-2 & SUBJ  & Sympt. & Yelp  \\ 
    \midrule
        w/o Coreset   & 35.32 & 64.24 & 71.94 & 54.51 & 89.32 & 73.10 & 77.23 & 53.52 & 87.42 & 95.19 & 87.58 & 55.87 \\
        w/o Selective & 35.94 & 66.40 & 73.35 & 52.14 & 89.14 & 75.46 & 77.23 & \underline{55.69} & 87.90 & 95.49 & 89.95 & 56.27 \\
        w/o Aug       & \underline{37.82} & 64.86 & 72.82 & 43.89 & 89.43 & 78.65 & 76.74 & \underline{55.69} & 86.75 & \underline{95.70} & 86.06 & 53.74 \\
        w/o DPO|DS    & 37.64 & \underline{70.18} & \underline{75.52} & 59.04 & \underline{89.72} & 79.90 & 78.42 & 54.87 & 87.99 & 95.47 & 90.08 & 56.50 \\
        w/o DPO       & 37.32 & 69.49 & 74.80 & 58.75 & 89.34 & \underline{80.67} & 78.43 & 55.37 & \underline{88.15} & 95.37 & \underline{90.66} & \underline{56.52} \\
        w/o DS        & 36.85 & 69.62 & 75.49 & \underline{59.32} & 89.53 & 80.18 & \underline{78.56} & 53.70 & 88.14 & 95.51 & 90.65 & 56.36 \\ 
    \midrule
        \Methodnameb~ & \textbf{38.46}  & \textbf{70.22}  & \textbf{75.62}  & \textbf{59.76}  & \textbf{89.78}  & \textbf{80.97}  & \textbf{80.10}  & \textbf{56.05}  & \textbf{88.64}  & \textbf{95.80}  & \textbf{90.74}  & \textbf{56.57}  \\ 
    \bottomrule 
    \end{tabular}
}
    \caption{Ablation study on model performance gains.}
    \label{tab:ablation}
\end{table}

\begin{table}[t]
    \centering
    \scriptsize
    \setlength{\tabcolsep}{4pt} 
    \renewcommand{\arraystretch}{1} 
    \begin{tabular}{lccccccc}
        \toprule
         & \multicolumn{3}{c}{GPT-2 (137M)} &  & \multicolumn{3}{c}{T5-large (738M)} \\ 
        \cmidrule{2-4} \cmidrule{6-8}
                            & CoLA  & MNLI  &  RCT  &  & CoLA  & MNLI  &  RCT  \\
        \midrule
        Original            & 67.64 & 50.03 & 76.53 &  & 78.94 & 50.06 & 78.98 \\
        OCR                 & 65.51 & 54.42 & 77.70 &  & 77.79 & 56.35 & 81.77 \\
        Keyboard            & 65.23 & 54.51 & 77.37 &  & 77.06 & 56.40 & 81.65 \\
        EDA                 & 66.60 & 54.13 & 76.89 &  & 78.59 & 51.58 & 80.79 \\
        ADEA                & 67.40 & 56.03 & 77.73 &  & 78.62 & 57.93 & 82.07 \\
        BT                  & 63.43 & 53.89 & 77.10 &  & 76.22 & 54.70 & 80.80 \\
        Unmask              & 66.87 & 55.75 & 78.01 &  & 77.84 & 61.88 & 81.41 \\
        AugGPT              & 66.08 & 54.80 & 78.28 &  & 77.79 & 60.18 & 79.18 \\
        Grammar             & 66.20 & 55.75 & 76.96 &  & 76.52 & 57.80 & 81.03 \\
        Spelling            & 66.24 & 55.53 & 78.16 &  & 77.57 & 62.93 & 81.99 \\
        Chain               & 65.85 & 54.83 & 78.04 &  & 77.98 & 59.88 & 81.50 \\
        Hint                & 66.41 & 55.75 & 77.52 &  & 78.06 & 61.13 & 81.38 \\
        Taboo               & 64.27 & 56.13 & 77.72 &  & 77.36 & 61.55 & 81.30 \\
        w/o Aug             & 67.85 & 52.79 & 79.47 &  & 79.74 & 53.18 & 81.63 \\
        w/o Coreset         & 65.24 & 50.89 & 76.98 &  & 77.92 & 51.85 & 81.01 \\
        w/o Selective       & 66.05 & 50.23 & 78.76 &  & 79.00 & 53.22 & 80.85 \\
        w/o DPO|DS          & 66.74 & 55.51 & 79.71 &  & 79.64 & 61.06 & 82.09 \\
        w/o DPO             & 67.05 & 54.97 & 79.72 &  & 79.51 & 61.79 & 81.83 \\
        w/o Sampling        & 67.50 & 55.93 & 79.79 &  & 78.97 & 62.03 & 82.18 \\
        \midrule
        \Methodnamea        & \textbf{68.14} & \textbf{56.25} & \textbf{79.83} &  & \textbf{79.93} & \textbf{63.91} & \textbf{82.20} \\
        \bottomrule
    \end{tabular}
    \captionof{table}{Training the GPT-2 and T5-large models on CoLA, MNLI, and RCT dataset.}
    \label{tab:t5}
\end{table}

\begin{figure}[t]
    \centering
    \subfloat[Accuracy]{
        \includegraphics[width=0.30\linewidth]{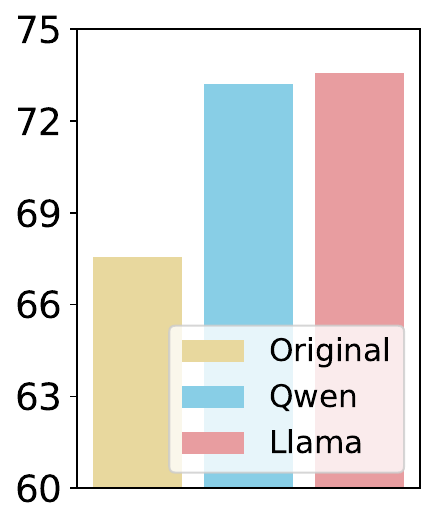}
        \label{subfig:qwen_perf}
    }
    \subfloat[Distance]{
        \includegraphics[width=0.31\linewidth]{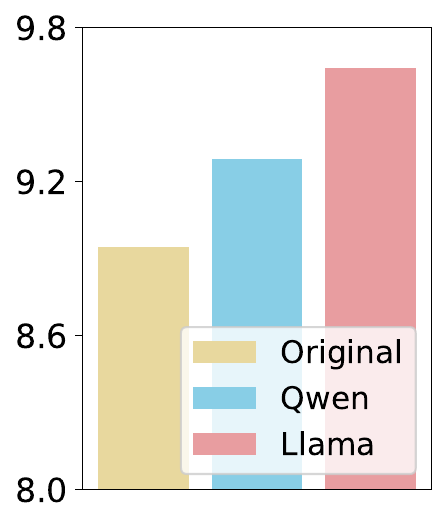}
        \label{subfig:qwen_dist}
    }
    \subfloat[3-grams]{
        \includegraphics[width=0.35\linewidth]{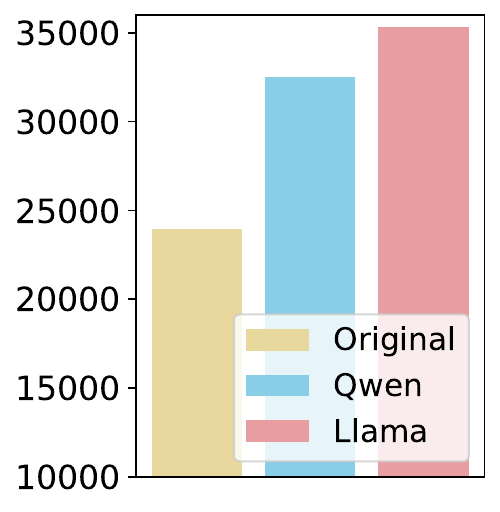}
        \label{subfig:qwen_disp}
    }
    \caption{Performance and diversity comparison between Llama and Qwen.}
    \label{fig:qwen}
\end{figure}

\subsection{LLM Architectures Adaptability}
To exhibit the generalizability of our proposed methodology, we replace the LLM augmenter and downstream task model with other LLM architectures respectively. The results show that \Methodnamec~is agnostic to LLM architectures.

For the LLM augmenter, we replace the Llama-3.2-1B-Instruct model with the similar-sized Qwen2.5-1.5B-Instruct model~\cite{qwen2.5}. 
As shown in Figure~\ref{fig:qwen}, datasets augmented by the Qwen model significantly outperform the original datasets in diversity and achieve comparable accuracy with those of the Llama model. 
Detailed results are given in Appendix~\ref{app:llm_arch}.

For the downstream task model, we replace \(\text{BERT}\), an encoder-only model with the GPT-based and T5-based classification models. 
GPT is a decoder-only LLM and is especially adept at generating texts from a prompt. Based on an encoder-decoder transformer architecture, T5 is trained to perform all NLP tasks in a unified text-to-text format and is favorable in broad cases. Specifically, we use GPT-2~\cite{radford2019language} and T5-large~\cite{raffel2020exploring} as the backbone of classification models, train these models on the MNLI, CoLA, and RCT datasets, and collect their performances. 
Experimental results in Table~\ref{tab:t5} show that \Methodnamec~ benefits both decoder-only models such as GPT and encoder-decoder models such as T5. 

\section{Conclusion}

Diversity is an important factor in developing AI-ready and high-quality datasets but is often ignored in data augmentation methods. 
We propose an innovative \textbf{\underline{D}}iversity-\textbf{\underline{o}}riented data \textbf{\underline{Aug}}mentation framework (\textbf{DoAug}) that fine-tunes an LLM paraphraser to enlarge and introduce diversity to textual datasets. 
The LLM paraphraser is fine-tuned to rewrite existing seed sentences in the original datasets, generating high-affinity samples and ensuring coherence of the dataset. 
We further construct a preference dataset and then fine-tune the LLM paraphraser with the DPO algorithm to encourage diversified generation. 
In this way, we maximize the diversity of the augmented dataset in our method. 
In extensive experiments, our proposed method exhibits a remarkable capability to boost dataset diversity, and the diversity gain significantly benefits the model's learning performance of downstream tasks.

\section{Acknowledgments}
Pengfei Wang is supported by the National Natural Science Foundation of China (Grant Nos. 62406306 and 92470204), and the Science and Technology Development Fund (FDCT), Macau SAR (file no. 0123/2023/RIA2, 001/2024/SKL).  

\section{Limitations}
This study has several limitations that should be acknowledged and addressed in future work.

\noindent\textbf{Diversity Exploration}: The evaluation of diversity lacks agreement on universally accepted metrics. 
In this study, we employed a subset of diversity-related evaluation methods, but other metrics, such as human-centered diversity evaluation, were not utilized. 
This limitation suggests that our assessment of diversity may not fully capture all aspects of the concept.

\noindent\textbf{Augmentation Validation}: Evaluating the correctness of generated data remains a challenging task. 
While both human evaluation and model-assisted evaluation are viable approaches, each comes with its own limitations. 
In this study, we employed both a task-aware LLM and humans for evaluation. 
However, there is no perfect human or LLM and this approach has inherent constraints, such as potential biases in the LLMs' training corpus or humans' knowledge and their inability to fully capture nuanced correctness in certain contexts.

\noindent\textbf{Generation Factors}: The quality and characteristics of generated samples are influenced by multiple factors, including the generation temperature, the choice of prompts, and the specific LLMs used. 
In this study, for each above-mentioned factor, we only explored two different settings, and we did not exhaustively explore all possible configurations. 
This restriction may have impacted the diversity and quality of the generated samples.

\noindent\textbf{Evaluation Benchmarks}: Our evaluation was primarily focused on sentence classification tasks and included two QA-based reasoning tasks, and we did not extend our analysis to more general tasks, such as mathematical reasoning, instruction-following, creative writing, or chain-of-thought (CoT) reasoning. 
Additionally, the datasets used in this study concentrated on English corpora, and we only considered one multilingual dataset, which could offer insights into cross-lingual or language-specific performance. 
Furthermore, we did not explore multimodality scenarios, which could provide a broader perspective on the applicability of our framework. 

\noindent\textbf{Potential Risks of Using LLM}: Leveraging LLMs for data augmentation might suffer from demographic bias and factual inaccuracies. First, LLMs may amplify demographic biases from their training data. Second, generation hallucinations may produce plausible but factually incorrect content. When task models train on such flawed data, their reliability and accuracy degrade, especially in high-stakes domains like healthcare or finance. Mitigating these risks requires rigorous validation, bias-detection frameworks, and human oversight to ensure the generated datasets uphold fairness and factual integrity.

These limitations highlight potential areas for future work, especially the adoption of more comprehensive diversity metrics and evaluation across diverse data modalities.

\bibliography{ref}

\clearpage
\appendix

\section{Evaluation Criterion}
$\bullet$ \textit{Distance} assesses the average distance between samples as follows:
    \({Distance(\mathcal{S}) = \frac{1}{|\mathcal{S}|} \sum_{x_i, x_j \in \mathcal{S}} \sqrt{(e_{x_i} - e_{x_j})^2}}\), 
where \(e_x = \mathcal{E}(x)\) is the embedding of sample \(x\) in the embedding space \(\mathcal{E}\), and a larger distance indicates greater diversity.  

$\bullet$ \textit{Dispersion}~\cite{yu2022can} is similar to cosine similarity but adjusted to make larger dispersion indicate greater diversity: 
        \(\mathit{Dispersion}(\mathcal{S}) = \frac{1}{|\mathcal{S}|} \sum_{x_i, x_j \in \mathcal{S}} 1 - \frac{e_{x_i} \cdot e_{x_j}}{\|e_{x_i}\|\|e_{x_j}\|}\). 

$\bullet$ \textit{Isocontour Radius}~\cite{lai2020diversity} 
is the geometric mean of the radii, reflecting the spread of embeddings along each axis. 
    Assuming sample embeddings follow a multivariate Gaussian distribution, the dataset can be taken as an ellipsoid-shaped cluster, formulated as: \(\sum_{j=1}^{H} \frac{(e_j - \mu_j)^2}{\sigma_j^2} = c^2\),
    where \(\mu_j\) is the embeddings' mean along the \(j\)-th axis, and \(\sigma_j^2\) is the variance of the \(j\)-th axis. 
    Geometrically, the standard deviation \(\sigma_j\), is the radius \(r_j\) of the ellipsoid along the \(j\)-th axis. Thus, we have: 
        \(\mathit{Radius}(\mathcal{S}) = (\prod_{i=1}^{H} \sigma_i)^{1/H}\). 

$\bullet$ \textit{Homogeneity}~\cite{lai2020diversity} is a metric that reflects the uniformity of a cluster distribution, suggesting that distinct samples in a diverse dataset should ideally cover the embedding space uniformly. 
    It begins by constructing a Markov chain model on the dataset embeddings. 
    The edge weight between sample \(i\) and \(j\) is defined as \(weight(i,j) = \left(\sqrt{(e_i - e_j) \cdot (e_i - e_j)}\right)^{\log{H}}\), and the transition probability from \(i\) to \(j\) is \(p(i \rightarrow j) = \frac{weight(i,j)}{\sum_{k} weight(i,k)}.\) 
    The entropy of the Markov chain is calculated by \(entropy(\mathcal{S}) = - \sum_{ij \in \mathcal{S}} v_i \cdot p(i \rightarrow j) \log{p(i \rightarrow j)}\), where \(v_i\) is the stationary distribution, assumed to be uniform. Homogeneity is then defined as, 
        \(\mathit{Homogeneity}(\mathcal{S}) = \frac{entropy(\mathcal{S})}{\log{(|\mathcal{S}|-1)}}\), 
    where \(\log{(|\mathcal{S}|-1)}\) is the entropy upper bound normalizes homogeneity into \([0,1]\)~\cite{lai2020diversity}. 
    
$\bullet$ \textit{Vocabulary Size} evaluates dataset diversity at the lexical level, complementing four embedding-level diversity metrics. Given the token set of the textual dataset \(\mathcal{T}\), we count the number of unique tokens present: 
        \(\mathit{Vocabulary}(\mathcal{S}) = |\mathcal{T}|\). 

$\bullet$ \textit{Unique 3-grams} is also a lexical level metric. By processing the textual dataset as a set of 3-grams \(\mathcal{G}_3\), we calculate its Unique 3-gram via:
    \(\mathit{3\text{-}gram}(\mathcal{S}) = |\mathcal{G}_3|\). 

The Distance, Dispersion, Isocontour Radius, and Homogeneity scores are calculated class-wise and then averaged over all classes, while vocabulary size and Unique 3-grams are directly derived from the entire dataset. Invalid (wrong) words are excluded when calculating lexical diversity. 

$\bullet$ \textit{Affinity} is defined as the reciprocal of the average deviation of class centers from the original dataset: 
$
    \mathit{Affinity}(\Tilde{S}, S) = \left(\frac{1}{|\mathcal{C}|} \sum_{c_i \in \mathcal{C}} \sqrt{(\Tilde{\mu}_{c_i} - \mu_{c_i})^2}\right)^{-1},
$
where \(\mathcal{C}={c_i}\) is the set of all classes, \(\Tilde{\mu}_{c_i}\) and \(\mu_{c_i}\) are the augmented and original embedding centers respectively.

\begin{table*}[htbp]
    \centering
    \small
    \renewcommand{\arraystretch}{1.2} %
    \begin{tabular}{ccccc}
        \toprule
         & Domain & Application task & Input & \#Classes \\
        \midrule
        ANLI~\cite{nie2020adversarial} & General & Entailment annotation & Sentence pair & 3 \\
        ChemProt~\cite{kringelum2016chemprot} & Medical & Chemical-protein relationship & Single sentence & 13 \\
        CoLA~\cite{wang2019glue} & General & Acceptability judgment & Single sentence & 2 \\
        MNLI~\cite{wang2019glue} & General & Entailment annotation & Sentence pair & 3 \\
        MPQA~\cite{wiebe2005annotating} & General & Sentiment analysis & Single sentence & 2 \\
        MRPC~\cite{wang2019glue} & General & Semantically equivalence & Sentence pair & 2 \\
        RCT~\cite{dernoncourt2017pubmed} & Medical & Role of sentence & Single sentence & 5 \\
        RTE~\cite{wang2019glue} & General & Entailment annotation & Sentence pair & 2 \\
        SST-2~\cite{wang2019glue} & General & Sentiment analysis & Single sentence & 2 \\
        SUBJ~\cite{pang2004sentimental} & General & Subjective v.s. objective & Single sentence & 2 \\
        Symptoms~\cite{dai2025auggpt} & Medical & Disease judgment & Single & 25 \\
        Yelp~\cite{zhang2015character} & General & Review rating & Single & 5 \\
        \bottomrule
    \end{tabular}
    \caption{A summary of 12 textual datasets.}
    \label{tab:datasets}
\end{table*}

\begin{table*}[htbp]
    \centering
    \small
    \renewcommand{\arraystretch}{1} %
    \scalebox{1}{
        \begin{tabular}{lp{6cm}cp{6cm}}
            \toprule
             & Symptoms & & SST-2 \\
            \midrule
            Original & My joints ache whenever it is cold & & I have always appreciated a smartly written motion picture \\
            \cmidrule{1-4}
            OCR & My joint5 aehe when€ver it is co1d && I hawo alwa9\$ appraciateol o smartly wri7tcn mo7ion pic+ure \\
            \cmidrule{1-4}
            Keyboard & My joibts axhe wjrnever it js cild && I have alwats apprecuated a smartly writren motion picthre \\
            \cmidrule{1-4}
            EDA & My joints ache whenever it is common cold & & I have always liked a smartly written motion picture\\
            \cmidrule{1-4}
            AEDA & My joints : ache whenever , it is ? cold & & I have . always appreciated ; a smartly written . motion picture \\
            \cmidrule{1-4}
            BT & My joint hurts when it's cold & & I always admire smart writing action pictures\\
            \cmidrule{1-4}
            Unmask & My joints ache. it is cold && I have always appreciated a smartly written motion picture \\
            \cmidrule{1-4}
            AugGPT & I experience joint pain during cold weather & & I've always been fond of motion pictures that showcase clever writing \\
            \cmidrule{1-4}
            Grammar & Whenever it is cold, my joints ache && A motion picture that has been always appreciated by me is smartly written \\
            \cmidrule{1-4}
            Spell & My joints ake whenever it is colld && I have allwas apreciated a smartly written motion picture \\
            \cmidrule{1-4}
            Chain & I have trouble with my joints in cold weather && I am a fan of movies that are skillfully made and have a captivating storyline \\
            \cmidrule{1-4}
            Hint & Cold weather often makes my joints feel stiff && I have a fondness for motion pictures that are well-written, well-crafted, and have a long-standing appreciation for them \\
            \cmidrule{1-4}
            Taboo & Whenever it gets chilly, my joints feel quite sore && I have a fondness for motion pictures that are well-crafted\\
            \cmidrule{1-4}
            \Methodnamec & I have trouble moving my joints in cold weather, causing discomfort & & I have a fondness for films that are well-crafted and have a sophisticated style \\
            \bottomrule
        \end{tabular}
    }
    \caption{Augmentation examples on the Symptoms and SST-2 datasets.}
    \label{tab:examples}
\end{table*}

\section{Dataset Specification}
\label{app:dataset}
The details of the 12 NLP datasets used in our experiments, including the domain, application task, input scheme, and class number, are summarized in Table~\ref{tab:datasets}. 
ANLI~\cite{nie2020adversarial}, MNLI~\cite{wang2019glue}, and RTE~\cite{wang2019glue} are three datasets that require models to recognize the textual entailment of two sentences, which requires decent reasoning ability of models. 
ChemProt~\cite{kringelum2016chemprot}, RCT~\cite{dernoncourt2017pubmed}, and Symptoms~\cite{dai2025auggpt} are three medical datasets that involve domain knowledge of the models. 
ChemProt describes the relationship between chemical-protein Paris, RCT requires the model to analyze what role a sentence plays in the abstract of a medical research paper, and the Symptoms dataset is about judging the disease from patient complaints. 
MPQA~\cite{wiebe2005annotating}, SST-2~\cite{wang2019glue}, and Yelp~\cite{zhang2015character} and sentiment datasets, where MPQA and SST-2 label the sentences as ``negative'' or ``positive'', and Yelp assigns numerical ratings from 1 to 5 to the reviews. 
CoLA~\cite{wang2019glue} evaluates the acceptability of a sentence, MRPC~\cite{wang2019glue} evaluates if a pair of sentences are equivalent, and SUBJ~\cite{pang2004sentimental} evaluates if a sentence is subjective or objective. 
Symptoms dataset is available at Kaggle\footnote{https://www.kaggle.com/datasets/paultimothymooney/medical-speech-transcription-and-intent}.
GLUE benchmark datasets (CoLA, MNLI, MRPC, RTE, and SST-2)\footnote{https://huggingface.co/datasets/nyu-mll/glue}, 
ANLI\footnotetext{https://huggingface.co/datasets/facebook/anli}, ChemProt\footnote{https://huggingface.co/datasets/AdaptLLM/ChemProt}, RCT\footnote{https://huggingface.co/datasets/AdaptLLM/RCT}, MPQA\footnote{https://huggingface.co/datasets/rahulsikder223/SentEval-MPQA}, SUBJ\footnote{https://huggingface.co/datasets/SetFit/subj}, and Yelp\footnote{https://huggingface.co/datasets/Yelp/yelp\_review\_full} are also available at Hugging Face. 

\section{Implementation Details} 
\label{app:implemtation}
We use Llama-3.2-1B-Instruct with BF16 quantization as the LLM paraphraser. 
Llama-3.2 is one of the latest products of the Llama family. The Llama-3.2-1B model outperforms the Llama-3.2-3B and Llama-3.1-8B models on the rewriting task while requiring minimal memory and inference time\footnote{https://huggingface.co/meta-llama/Llama-3.2-1B-Instruct}. 
The prompt for paraphrasing is ``You will be given a sentence. Please paraphrase the sentence.'' 
We fine-tune the Llama model with the LlamaFactory framework~\cite{zheng2024llamafactory}. 
When training the model, we use LoRA to reduce computation costs. 
We train a LoRA adapter for the SFT stage, merge the SFT adapter, train a LoRA adapter for the DPO stage, and finally merge the DPO adapter for use. 
In the SFT stage, the learning rate is \(10^{-4}\). In the DPO stage, \(\beta\) in the loss function is set to \(0.1\) and the learning rate is \(5^{-6}\). 
The LLM is trained for 3 epochs with the AdamW optimizer, a cosine scheduler, and a warm-up ratio of \(0.1\) in each stage. 
The rank \(r\) for LoRA fine-tuning is \(8\). 
\(\mathcal{D}_{\text{SFT}}\) contains 100,000 sentence pairs (20,000 original sentences and 5 paraphrases for each of them). 
In the SFT stage, the model is trained to produce one paraphrase for one input. 
\(\mathcal{D}_{\text{DPO}}\) contains 50,000 preference pairs. 
We use the embedding vector of the \texttt{[CLS]} token in the last layer of the \(\text{BERT}_{\text{base}}\) model as the embedding space \(\mathcal{E}\) for both preference dataset construction and dataset diversity evaluation. 
In the coreset selection step, the coreset ratio \(r_\text{augment} : r_\text{retain} : r_\text{prune}\) is \(1:1:1\). 
For each original sentence, we generate \(K=5\) paraphrases and sample the most diversified output. 
For downstream task evaluation, we train the BERT model for 3 epochs. 
The model is updated with the AdamW optimizer. The learning rate is \(5^{-5}\) with a linear scheduler and no warm-up. 
Experiments are repeated with ten random seeds. 
The LLM paraphraser and downstream task models are trained on two A100-40G GPUs with \texttt{transformers} 4.45.2, \texttt{pytorch} 2.5.1, and CUDA 12.4. 
Training with SFT and DPO takes 32 and 36 minutes, respectively, and the LLM augmenter can paraphrase roughly one sentence per second.

\section{Baseline Methods Implementation}
The error ratios for OCR and Keyboard are set to \(0.15\).
The ratios for EDA's four operations are set to \(0.1\). 
AEDA's punctuation ratio is \(0.3\) as used in the original method. 
For BT, we use the en-zh and zh-en versions of the opus-mt model by Helsinki-NLP~\cite{TiedemannThottingal:EAMT2020} as the translator. 
The masking ratio for Unmask is set to \(0.15\) and we sample the top-1 predictions of the \(\text{BERT}_{\text{base}}\) model. 
Considering the cost of calling ChatGPT APIs, when implementing AugGPT, Grammar, Spelling, Chain, Hint, and Taboo, we use the open-source Llama-3.1-8B-Instruct model~\cite{dubey2024llama} as a replacement. Since it achieves competitive performance compared with the GPT 3.5 Turbo model (e.g. 69.4 v.s. 70.7 on MMLU 5-shot and 80.4 v.s. 69.9 on IFEval)~\cite{dubey2024llama}, replacing GPT with Llama hardly compromises the effectiveness of the baseline methods. 
For the original dataset and all baselines, we also set the coreset ratio \(r_\text{augment} : r_\text{retain} : r_\text{prune}\) to \(1:1:1\) but do not rank the samples before selecting; for the original dataset, we do not augment \(\mathcal{S}_{\text{augment}}\). 
In this way, the number of samples from the original dataset is the same for all methods, ensuring fairness in terms of the samples' coverage and distribution.
The number of final samples is in accordance with DoAug for all baselines, further ensuring the fairness of model training and evaluation.
We use the same random seeds for all baselines as for ours. 

\section{Augmentation Examples}
\label{app:example}

We include some augmentation examples of \Methodnamec~and baseline methods in Table~\ref{tab:examples}. We can observe that \Methodnamec~introduces more details (have trouble moving) and some novel vocabularies (fondness, sophisticated).

\section{Diversity Evaluation}
\label{app:diversity}
We present the original diversity scores of all methods in Table~\ref{tab:diversity}.

\begin{table}[htbp]
    \centering
    \tiny
    \setlength{\tabcolsep}{3pt} 
    \begin{tabular}{lcccccc}
        \toprule
              &   {Distance }   &   {Dispersion }    &   {Radius }      &   {Homogeneity }     &   {Vocabulary } 
              & {\makecell[c]{3-grams} } \\

        \midrule

Original & 8.9440 & 0.1902 & 0.5354 & 0.8931 & 4734 & 23954 \\

OCR & 8.9514 & 0.1915 & 0.5340 & 0.8992 & 4793 & 25195 \\
Keyboard & 8.9474 & 0.1915 & 0.5334 & 0.8993 & 4829 & 25844 \\
EDA & 9.1324 & 0.2024 & 0.5287 & 0.9015 & 4974 & 29200 \\
ADEA & 8.9546 & 0.1926 & 0.5302 & 0.9073 & 4735 & 25596 \\
BT & 9.2111 & 0.2018 & 0.5347 & 0.8798 & 5193 & 30634 \\
Unmask & 9.0042 & 0.1932 & 0.5346 & 0.8994 & 4797 & 26774 \\
AugGPT & 9.1025 & 0.1956 & 0.5366 & 0.8757 & 5035 & 27420 \\
Grammar & 9.3905 & 0.2073 & 0.5373 & 0.8441 & 4895 & 30108 \\
Spelling & 9.0390 & 0.1949 & 0.5327 & 0.8996 & 5352 & 28154 \\
Chain & 9.1346 & 0.1955 & 0.5371 & 0.9076 & 5341 & 31543 \\
Hint & 9.3326 & 0.2043 & 0.5371 & 0.9016 & 5296 & 31649 \\
Taboo & 9.1241 & 0.1952 & 0.5368 & 0.9107 & 5172 & 30634 \\

\midrule
\textbf{DoAug} & 9.6430 & 0.2180 & 0.5362 & 0.9095 & 5993 & 35308 \\

        \bottomrule 
    \end{tabular}
    \caption{6 diversity metrics averaged on 12 datasets.}
    
    \label{tab:diversity}
\end{table}

\section{Measurements for Preference Dataset Construction and Diversity Sampling}

\begin{figure}[htbp]
    \centering
    \subfloat[Correlation]{
        \includegraphics[width=0.45\linewidth]{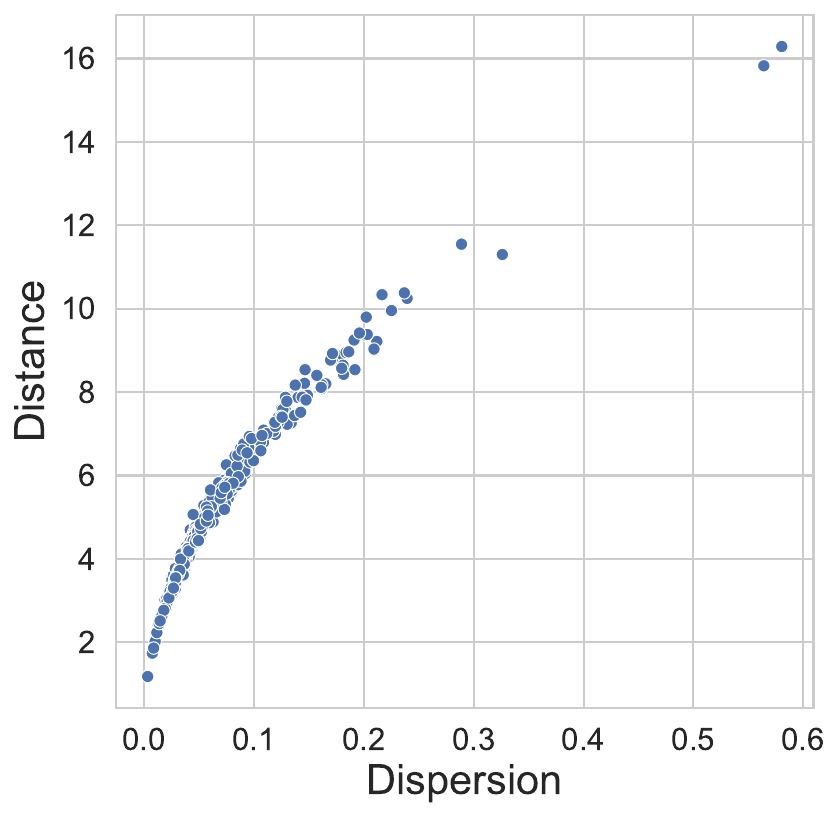}
        \label{subfig:disp_dist}
    }
    \subfloat[Distribution]{
        \includegraphics[width=0.45\linewidth]{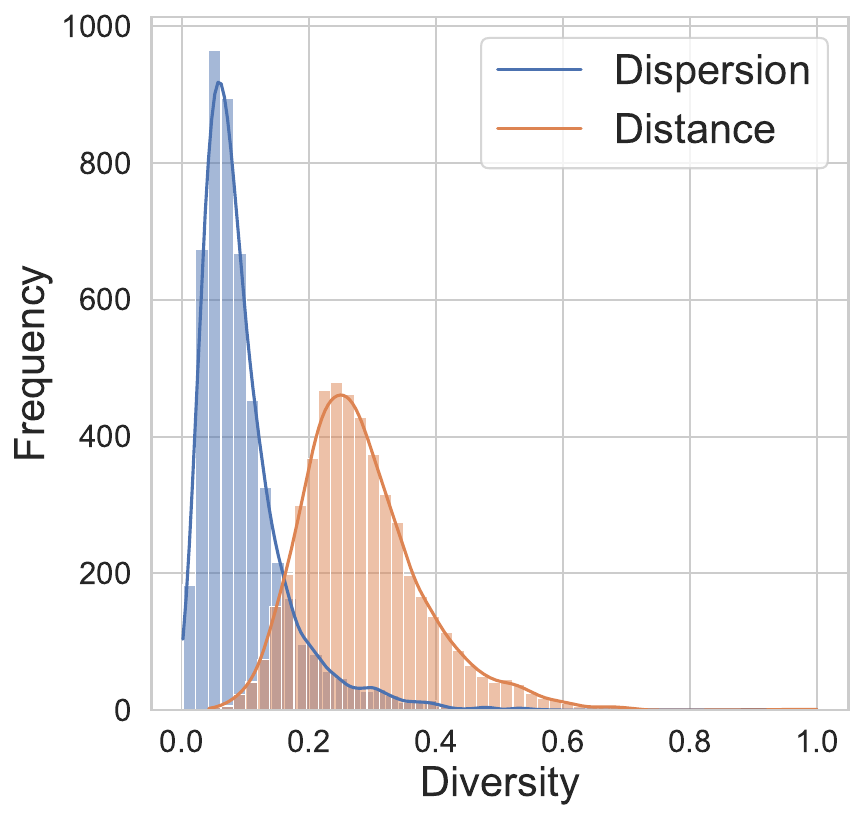}
        \label{subfig:disp_dist_distribution}
    }
    \caption{ For (\ref{subfig:disp_dist}), we sample 500 points when plotting the diagram. For (\ref{subfig:disp_dist_distribution}), all scores are normalized to \([0,1]\).}
    \label{fig:disp_dist}
\end{figure}

In our final experimental settings, we use Euclidean distance to construct the preference dataset and sample the more diversified generations. 
An alternative to Euclidean distance is Cosine similarity. Before we settled on Euclidean distance, we examined and compared both Euclidean distance and Cosine similarity. 
We conduct the examination on a subset of 1000 samples from the original paraphrase dataset. 
We notice that when selecting the most diverse samples as ``chosen samples'', in \(97.1\%\) of the cases the two metrics yield the same samples. 
When selecting the most repetitive samples as ``rejected samples'', in \(97.8\%\) of the cases, the two metrics yield the same samples.
For the total 5000 paraphrases (each sample contains 5 paraphrases), we also investigate the Pearson correlation between their Euclidean distance and dissimilarity (that is, 1 - Cosine similarity) compared with the original sentence, and find that the Pearson correlation is 0.96, indicating they are highly correlated. 
The relationship between Euclidean distance and dissimilarity for each paraphrase is shown in Figure~\ref{subfig:disp_dist}. 
Further, we observe that the sample distance is distributed more smoothly, suggesting samples of different diversity are more distinguishable, as shown in Figure~\ref{subfig:disp_dist_distribution}. 
Besides, Euclidean distance is more straightforward (higher distance is higher diversity) and easier to understand. 
Given the above arguments, we finally use Euclidean distance in our coding but expect the performance to be consistent if switching to Cosine similarity. 

\section{Full Results of Replacing DPO training}
\label{app:promptemp_app}
\begin{figure}[thbp]
    \centering
    \subfloat[Distance]{
        \includegraphics[width=0.14\textwidth]{diversity_prompt_temp_Distance.pdf}
        \label{subfig:prompt_temp_dist_app}
    }
    \subfloat[Dispersion]{
        \includegraphics[width=0.145\textwidth]{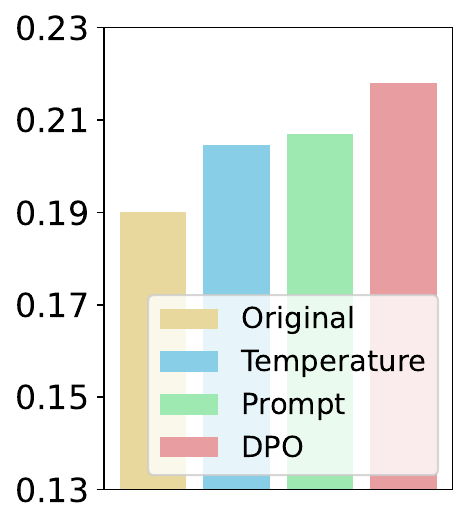}
        \label{subfig:prompt_temp_disp_app}
    }
    \subfloat[Radius]{
        \includegraphics[width=0.151\textwidth]{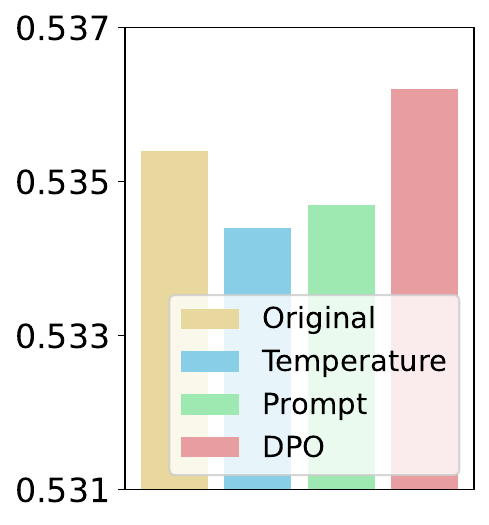}
        \label{subfig:prompt_temp_rad_app}
    }
    \\
    \subfloat[Homogeneity]{
        \includegraphics[width=0.145\textwidth]{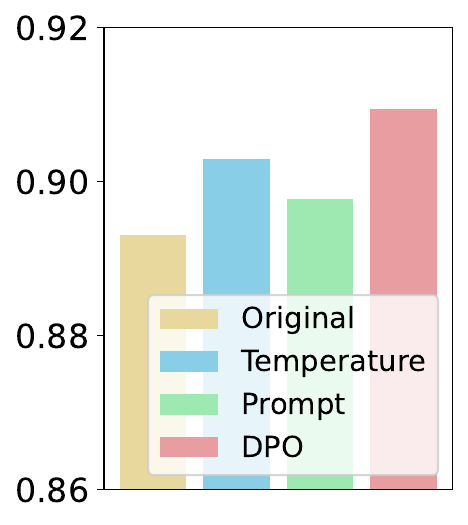}
        \label{subfig:prompt_temp_hom_app}
    }
    \subfloat[Vocabulary]{
        \includegraphics[width=0.147\textwidth]{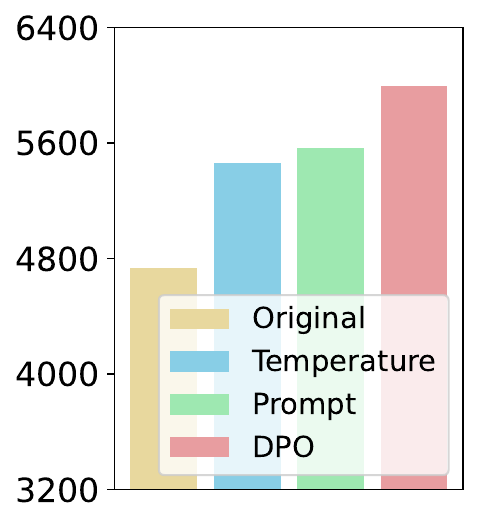}
        \label{subfig:prompt_temp_voc_app}
    }
    \subfloat[3-gram]{
        \includegraphics[width=0.151\textwidth]{diversity_prompt_temp_3-grams.pdf}
        \label{subfig:prompt_temp_3gram_app}
    }
    \caption{Replacement study on DPO training.}
    \label{fig:promptemp_diveristy_6}
\end{figure}

Detailed results of diversity in terms of 6 metrics are given in Figure~\ref{fig:promptemp_diveristy_6}.

\section{Parameter Sensitivity Study}

Given the heavy time and computation cost of training LLMs, we follow most settings from existing research or default configurations from the library. Still, we conduct sensitivity studies on some key parameters unique to our method.

\subsection{Number of Generated Sentences} 
First, we study the effect of \(K\), the number of total sentences generated by the LLM paraphraser when paraphrasing an original sentence. As Figure~\ref{subfig:num_seq} shows, the best performances are achieved when \(K\) is between \(5\) to \(8\). Intuitively, too small \(K\) limits the possibility and diversity of generated sentences and therefore affects the dataset diversity and task performance; on the other hand, too large \(K\) is likely to allow the LLM paraphraser go too far from the original semantics, breaking label preservability.

\subsection{Coreset Ratio} 
\label{app:ratio}
\begin{figure*}[htbp]
    \centering
    \vspace{-5mm}
    \subfloat[Generated sentences]{
        \includegraphics[width=0.36\linewidth]{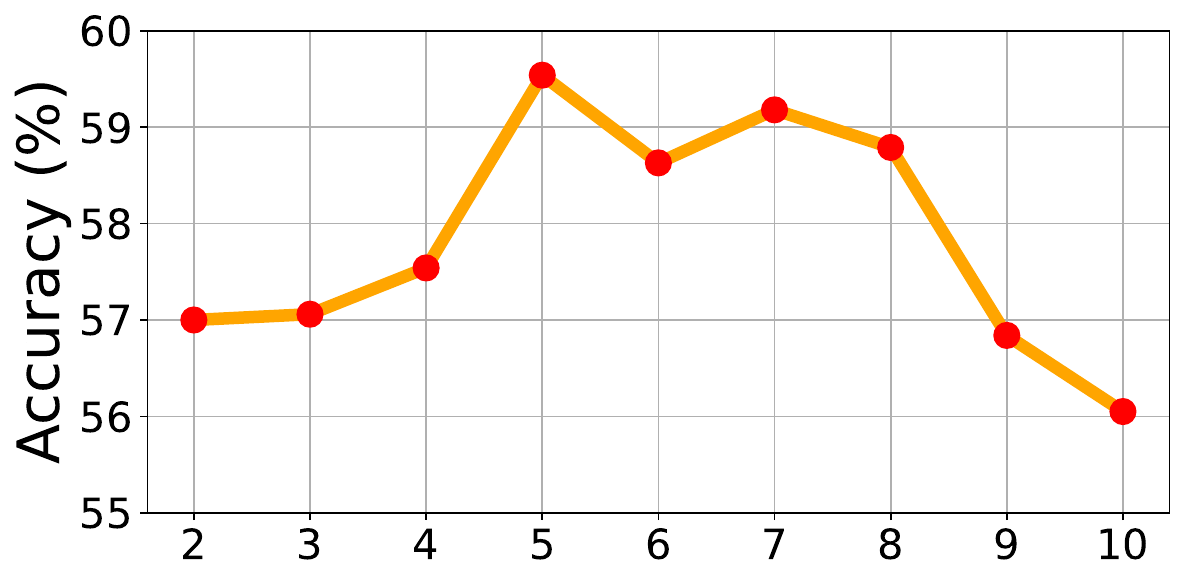}
        \label{subfig:num_seq}
    }
    \subfloat[Pruning ratio]{
        \includegraphics[width=0.27\linewidth]{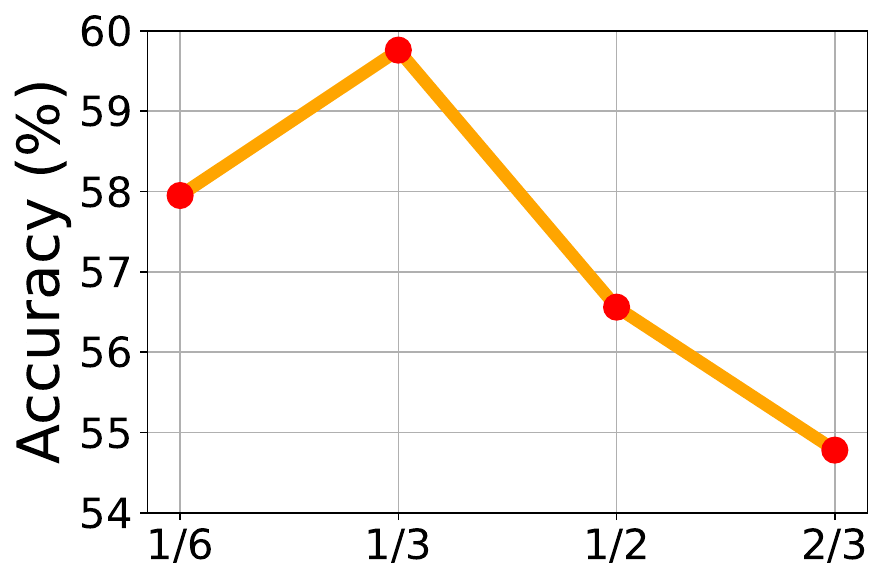}
        \label{subfig:dropselect}
    }
    \subfloat[Augmentation ratio]{
        \includegraphics[width=0.27\linewidth]{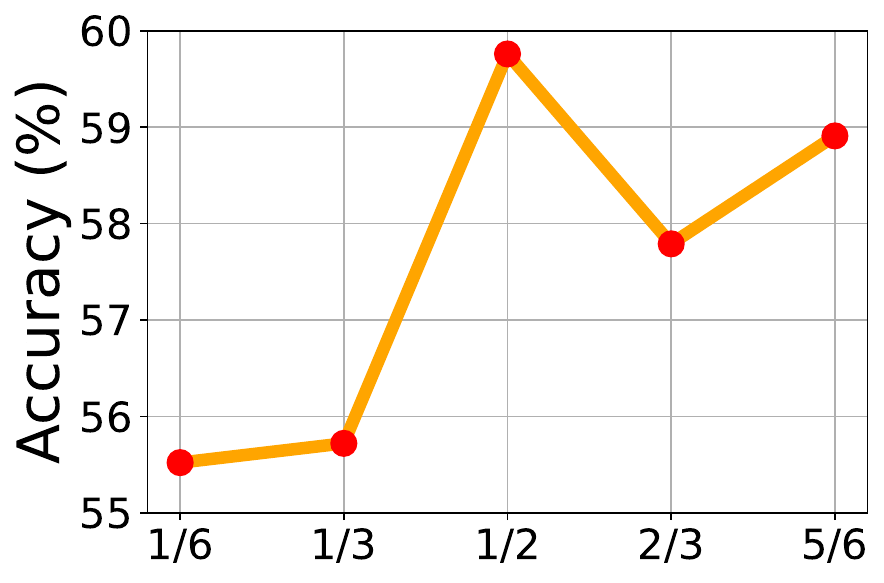}
        \label{subfig:augmentselect}
    }
    \caption{Parameter sensitivity study on generated sentence count, coreset ratio for data pruning and augmentation on MNLI dataset. Data pruning ratio is embodied by \(r_\text{prune} / (r_\text{prune} + r_\text{retain} + r_\text{augment})\), and data augmentation is embodied by \(r_\text{augment} / (r_\text{retain} + r_\text{augment})\).}
    \label{fig:sensitivity}
\end{figure*}

We study the effect of the coreset ratio for data pruning and data augmentation to determine the best ratios. 
As presented in Figure~\ref{subfig:dropselect} and Figure~\ref{subfig:augmentselect}, both data pruning and data augmentation favor a moderate ratio. 
The best performance occurs when the pruning ratio is \(1/3\) and the augmentation ratio is \(1/2\). 
In this way, \(1/3\) of the total samples are pruned, and \(2/3\) of the total samples are preserved.
Then from the preserved \(2/3\), we use \(1/2\) of the remaining as seed samples for augmentation, which is \((2/3) \times (1/2) = 1/3\) of the total samples. As a result, the final ratio \(r_\text{augment} : r_\text{retain} : r_\text{prune} = 1/3:1/3:1/3 = 1:1:1\).

\section{Coreset Selection Methods}
\begin{figure}[htbp]
    \centering
    \subfloat[CoLA]{
        \includegraphics[width=0.48\linewidth]{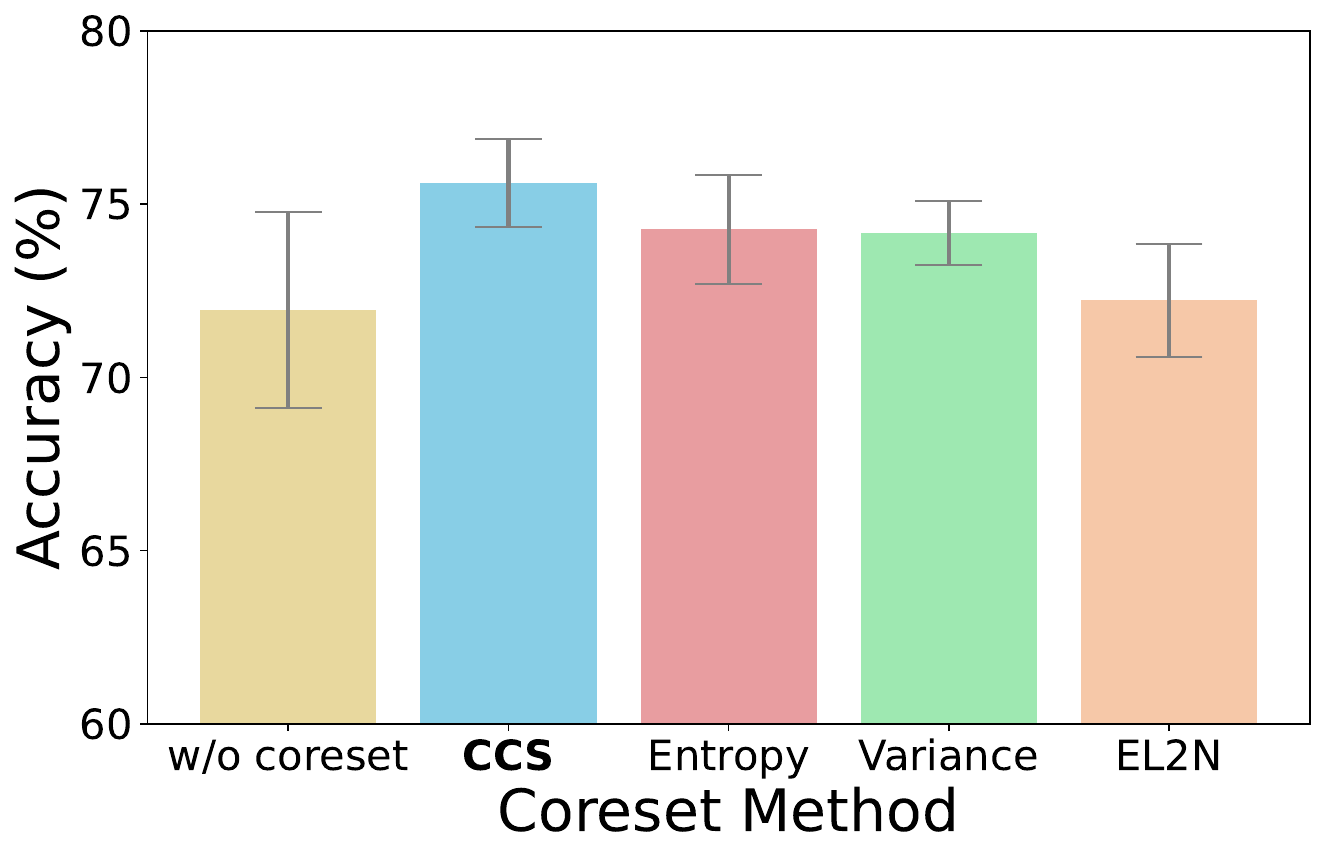}
        \label{subfig:sen_core_cola}
    }
    \subfloat[RCT]{
        \includegraphics[width=0.48\linewidth]{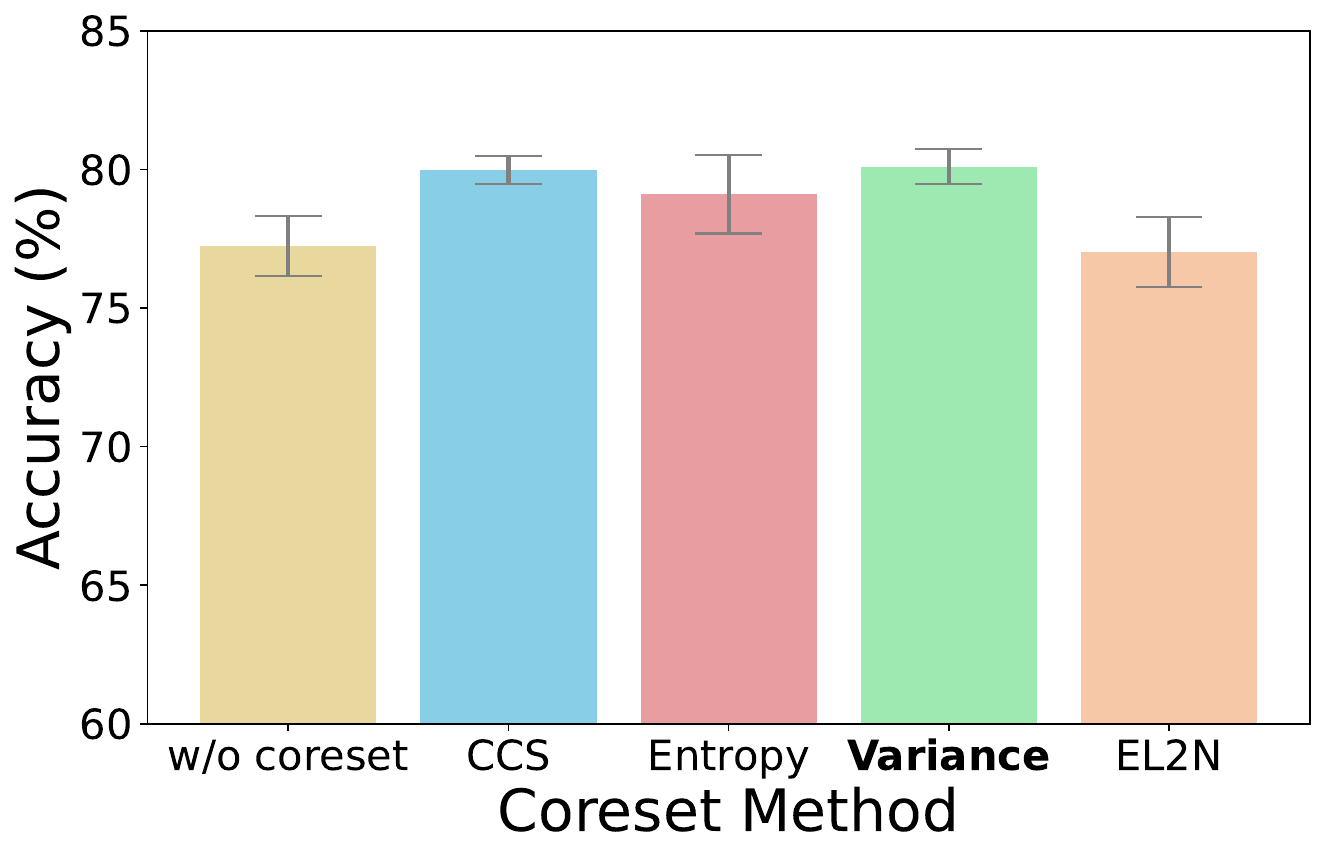}
        \label{subfig:sen_core_rct}
    }
    \\
    \subfloat[SST-2]{
        \includegraphics[width=0.48\linewidth]{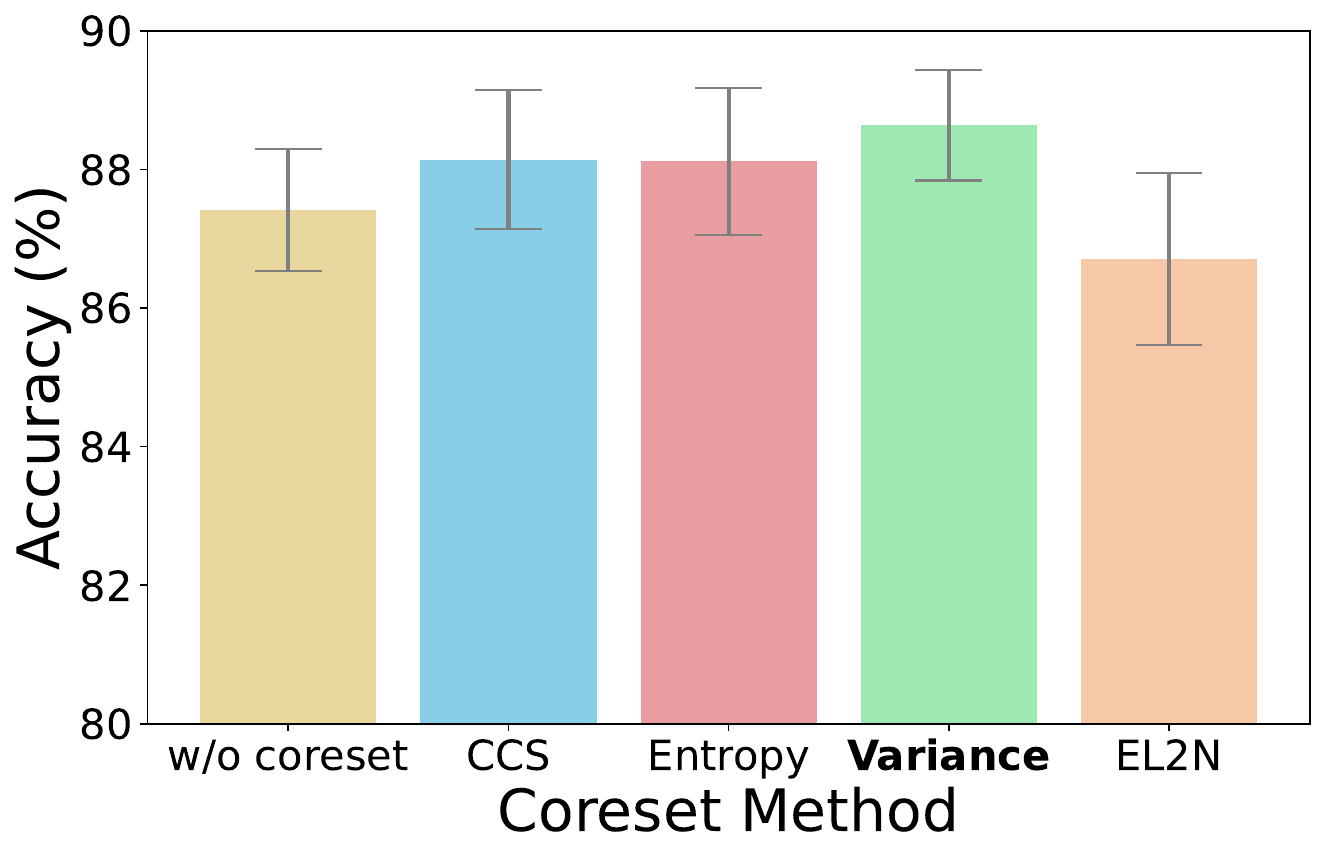}
        \label{subfig:sen_core_sst2}
    }
    \subfloat[SUBJ]{
        \includegraphics[width=0.48\linewidth]{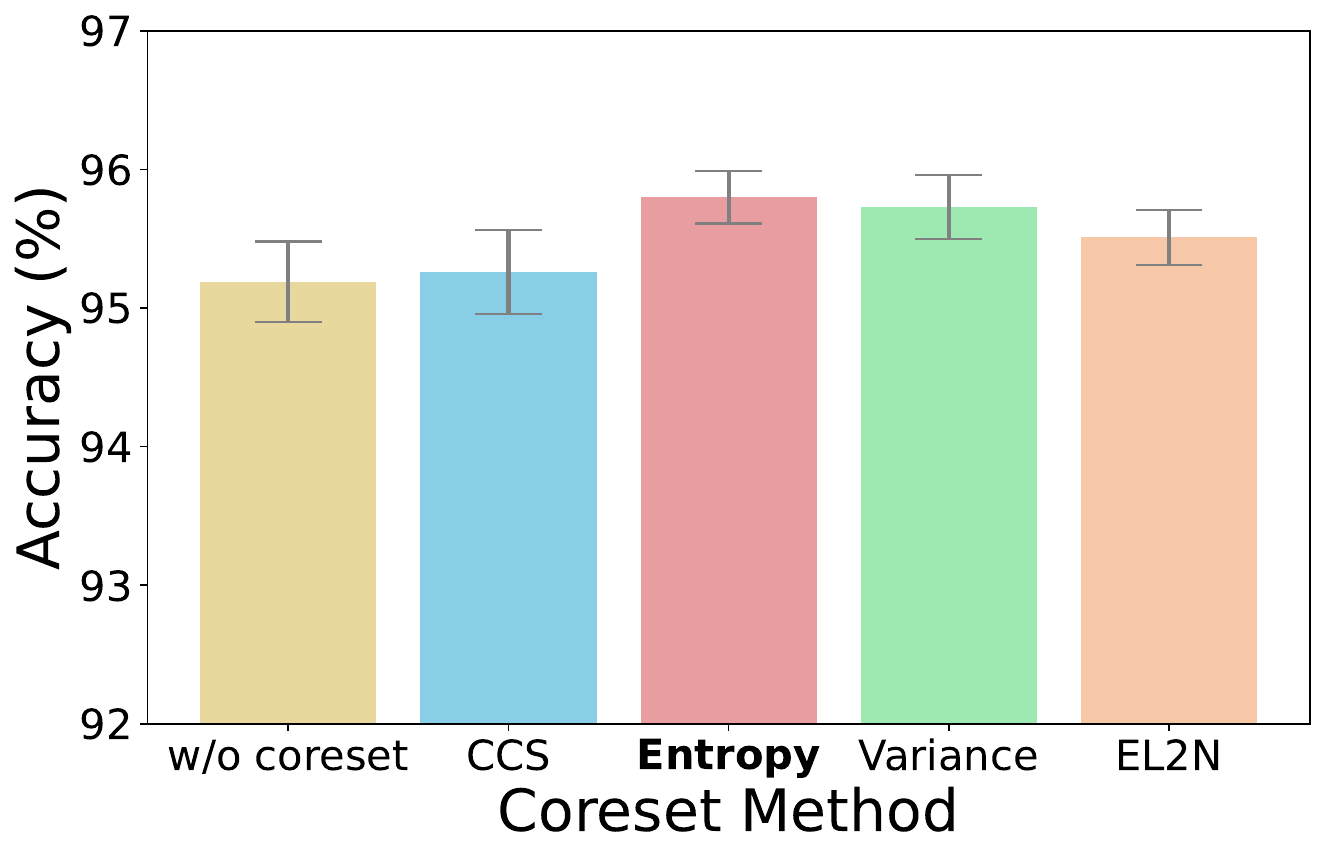}
        \label{subfig:sen_core_subj}
    }
    \caption{Sensitivity study on the choice of coreset method.}
    \label{fig:coreset_sensitivity}
\end{figure}

The choice of coreset methods is another factor that influences the final performance. 
So, we also investigate how the performance changes when augmentation is applied to different coresets, as presented in Figure~\ref{fig:coreset_sensitivity}.
The result shows that different datasets favor different coreset methods, however, we notice that ``variance'' and ``CCS w/ AUM'' benefit most datasets (9 out of 12), and in most cases, the suboptimal choices of coresets still outperform data augmentation without targeting coresets, and augmentation performance does not significantly degenerate on most suboptimal coresets. 
This result demonstrates our coreset-focused selective data augmentation method can benefit from appropriate coresets but is robust against suboptimal coresets, and the ``variance'' and ``CCS w/ AUM'' can be used as the default coreset method.

\section{Alleviating the Low-resource Problem}

In our main experiments, we artificially create low-resource conditions by sampling a subset from the original dataset. In Table~\ref{tab:low_res}, we compare the results of \Methodnamec~against model performance on large subsets, which are two times the size of that used in our main experiments. 
The comparison shows that \Methodnamec~can alleviate the low-resource problem and even performs better than larger subsets in some cases. 

\begin{table}[htbp]
    \centering
    \setlength{\tabcolsep}{1.8pt} 
    \tiny
    \begin{tabular}{cccccccccc}
    \toprule
     & Ch.Pr. & CoLA & MNLI & RCT & SST-2 & SUBJ & Sympt. & YELP \\ 
    \midrule
    800 Original Samples & 58.33 & 74.56 & 42.81 & 71.62 & 86.97 & 95.73 & 74.06 & 51.48 \\ 
    1.2 K Original  Samples & \textbf{71.12} & \textbf{77.61} & \textbf{61.63} & 78.96 & 88.56 & 95.77 & \textbf{92.45} & 56.48 \\
    \cmidrule{1-9}
    \Methodnamec & 70.22 & 75.62 & 59.76 & \textbf{80.10} & \textbf{88.64} & \textbf{95.80} & 90.74 & \textbf{56.57} \\
    \bottomrule
    \end{tabular}
    \captionof{table}{Results on low-resource datasets (800), larger original subsets (1.2 K), and~\Methodnamec~(800 Original + 400 Augmented Samples).}
    \label{tab:low_res}

\end{table}

\section{Full Results of LLM Architecture Exploration}
\label{app:llm_arch}

\begin{figure}[htbp]
    \centering
    \subfloat[Distance]{
        \includegraphics[width=0.14\textwidth]{diversity_qwen_Distance.pdf}
        \label{subfig:qwen_dist_app}
    }
    \subfloat[Dispersion]{
        \includegraphics[width=0.145\textwidth]{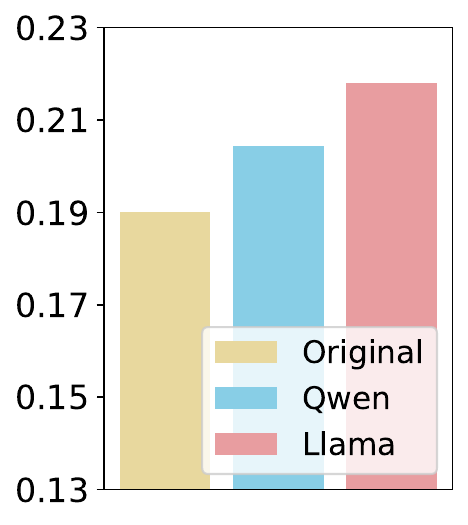}
        \label{subfig:qwen_disp_app}
    }
    \subfloat[Radius]{
        \includegraphics[width=0.151\textwidth]{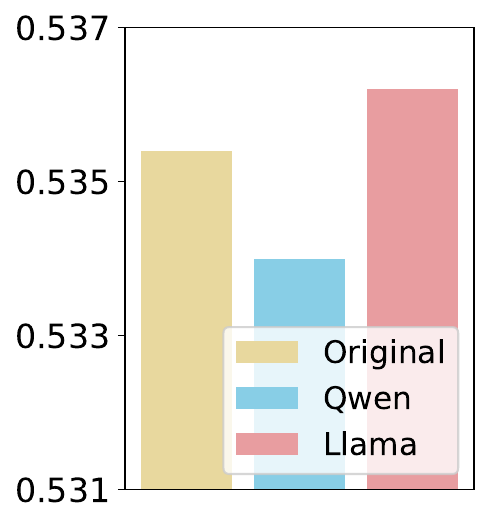}
        \label{subfig:qwen_rad_app}
    }
    \\
    \subfloat[Homogeneity]{
        \includegraphics[width=0.145\textwidth]{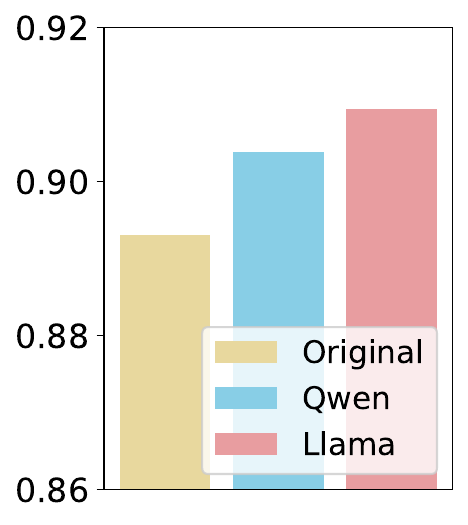}
        \label{subfig:qwen_hom_app}
    }
    \subfloat[Vocabulary]{
        \includegraphics[width=0.147\textwidth]{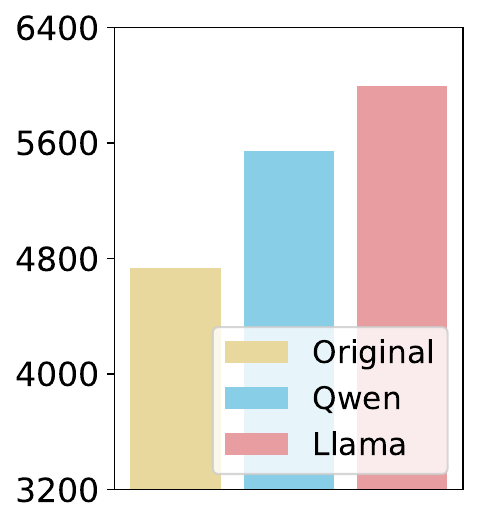}
        \label{subfig:qwen_voc_app}
    }
    \subfloat[3-gram]{
        \includegraphics[width=0.151\textwidth]{diversity_qwen_3-grams.pdf}
        \label{subfig:qwen_3gram_app}
    }
    \caption{Diversity comparison between Llama and Qwen}
    \label{fig:qwen_diveristy_5}
\end{figure}

\begin{figure*}[tbp!]
    \centering
    \includegraphics[width=.90\linewidth]{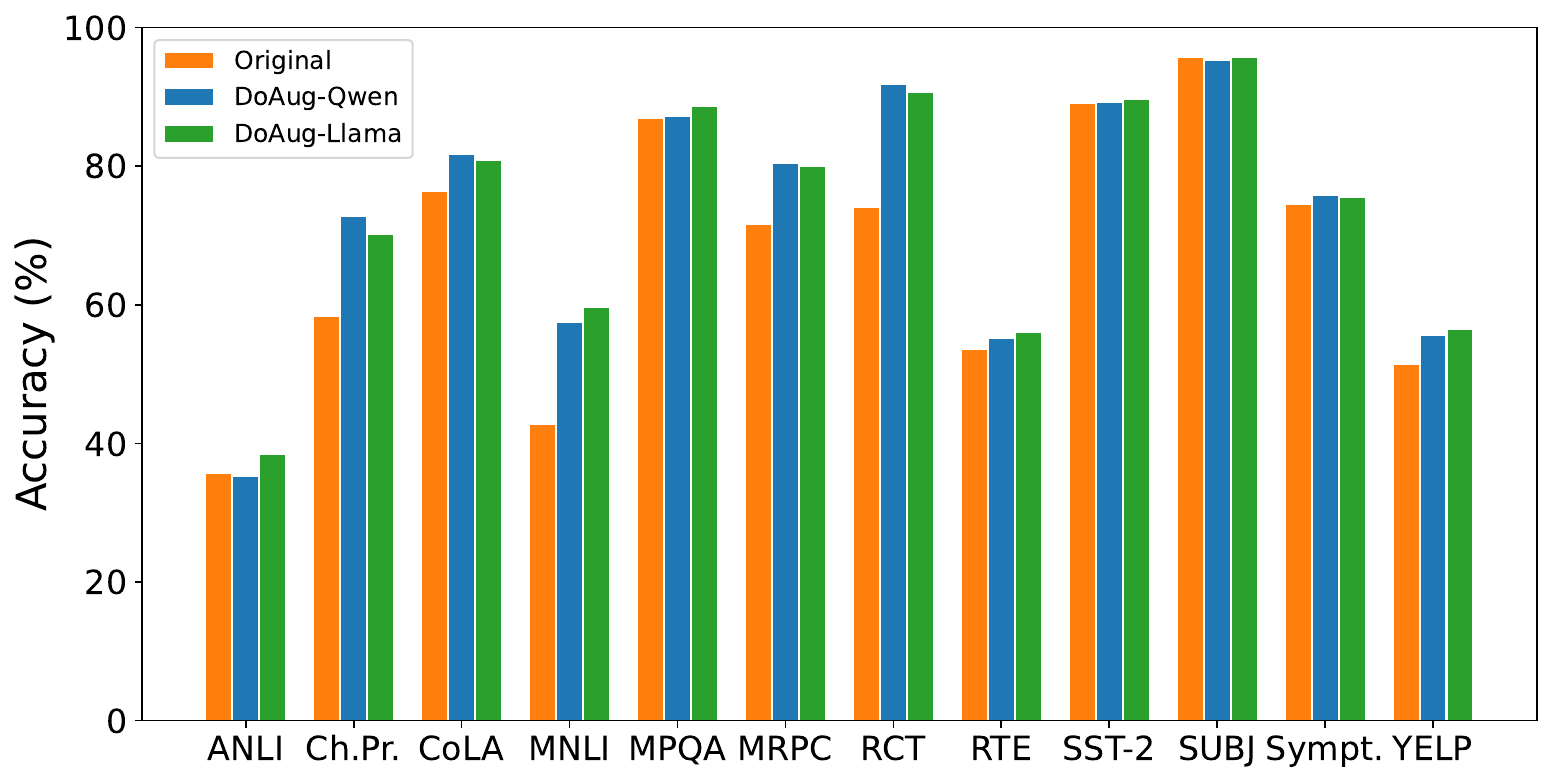}
    \caption{Qwen performance on 12 datasets}
    \label{fig:qwen_12}
\end{figure*}

Detailed results of performance on all 12 datasets and diversity in terms of 5 metrics are given in Figure~\ref{fig:qwen_12} and Figure~\ref{fig:qwen_diveristy_5}.

\begin{figure}[!t]
    \centering
    \subfloat[CSQA]{
        \includegraphics[width=0.3\linewidth]{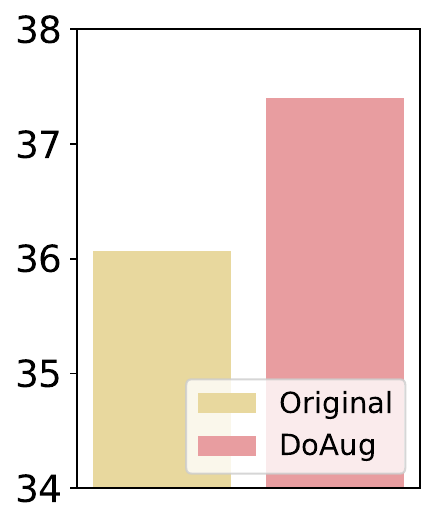}
        \label{subfig:csqa}
    }
    \subfloat[CODAH]{
        \includegraphics[width=0.3\linewidth]{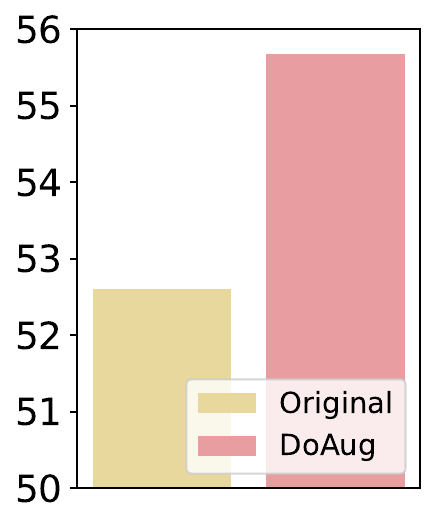}
        \label{subfig:codah}
    }
    \subfloat[Multilingual]{
        \includegraphics[width=0.3\linewidth]{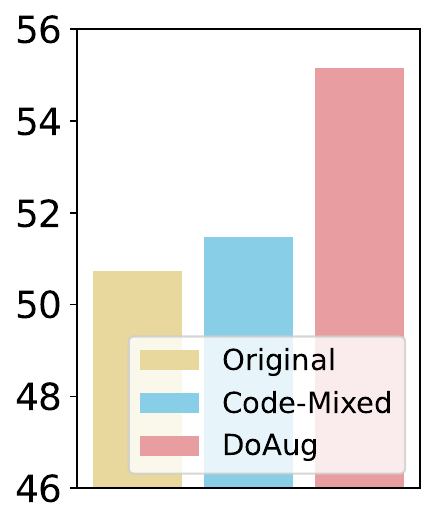}
        \label{subfig:multi}
    }
    \caption{Performance improvement on two multi-choice-formed reasoning datasets and a multilingual text classification dataset.}
    \label{fig:multi_reasoning}
\end{figure}

\section{Applicability on More Datasets}

To expand the evaluation scope and verify the broad application of \textbf{DoAug}, we additionally test the method on three datasets: CSQA, CODAH, and Multilingual, beyond the 12 English classification datasets in our main experiments. 
CSQA (CommonSenseQA)~\cite{talmor-etal-2019-commonsenseqa} and CODAH (COmmonsense Dataset Adversarially-authored by Humans)~\cite{chen-etal-2019-codah} are two reasoning datasets. They are both in the form of multiple-choice questions. 
Multilingual~\cite{muennighoff2023mteb,barbieri2022xlm,enevoldsen2025mmteb} is a multilingual sentiment analysis dataset collected from Twitter. We filter the dataset to keep English, Spanish, German, French, Italian, and Portuguese samples, and remove Arabic samples because our augmenter is a Llama model and does not support Arabic. 
For the Multilingual dataset, we also include Code-Mixed, a data augmentation technique that switches some words to other languages and is designed especially for multilingual sentiment analysis tasks. 
As shown in Figure~\ref{fig:multi_reasoning}, \textbf{DoAug} also benefits multilingual datasets and reasoning tasks.

\end{document}